%% file: main.tex
\documentclass[12pt]{article}

\PassOptionsToPackage{nopatch=footnote}{microtype}

\usepackage[nonatbib, preprint]{souls2014}

\input{packages}

\title{A Tutorial on Physics Constrained Machine Learning for Hybrid Modeling, Optimization, and Control}

\author{
  Rahul Golder \\Artie McFerrin Department of Chemical Engineering\\Texas A\&M University, College Station, TX 77843-3122, USA \\
  \texttt{rahulgolder8420@tamu.edu}
  \And
  Bimol Nath Roy \\Artie McFerrin Department of Chemical Engineering\\Texas A\&M University, College Station, TX 77843-3122, USA \\
  \texttt{bimolnathroy@tamu.edu}
  \And
  M. M. Faruque Hasan \thanks{Corresponding author: \texttt{hasan@tamu.edu}}\\Artie McFerrin Department of Chemical Engineering\\Texas A\&M University, College Station, TX 77843-3122, USA \\ Texas A\&M Energy Institute, Texas A\&M University, College Station, TX 77843, USA\\
  \texttt{hasan@tamu.edu} \\
}

\begin{document}

\maketitle

\begin{abstract}
Physics-constrained machine learning (PCML) integrates domain knowledge, first-principles, and physical constraints into AI/ML models. In this work, we present a step-by-step tutorial of a modular PCML framework, \texttt{KKT-HardNet}, and illustrate how it could be used for hybrid modeling, process simulation, optimization and model predictive control in a single platform. KKT-HardNet combines fast inference capabilities of neural networks with guaranteed satisfaction of algebraic constraints through differential projection. It can handle heterogeneous data and enforce both linear and nonlinear equality and inequality constraints. The modular design supports custom neural-network backbones and user-defined projection mechanisms for PCML. To enable efficient end-to-end training, KKT-HardNet exploits implicit function theorem to backpropagate through the projection layer. This avoids computationally expensive automatic differentiation. This work also introduces an exact reduced-order Fischer--Burmeister  reformulation and a primal--dual interior-point method as an alternative projection solve strategy. We also provide a diverse set of benchmarking problems for PCML in engineering applications. KKT-HardNet achieves orders of magnitude better performance in terms of both constraint satisfaction and inference time. The KKT-HardNet code and the benchmark datasets are available at \href{https://github.com/souls-tamu/kkt-hardnet}{\texttt{https://github.com/souls-tamu/kkt-hardnet}}.
\end{abstract}

\faGithub\,\,\, \href{https://github.com/souls-tamu/kkt-hardnet}{\texttt{https://github.com/souls-tamu/kkt-hardnet}} \hfill \faPython\,\,\, \href{https://pypi.org/project/kkt-hardnet/}{\texttt{pip install kkt-hardnet}}



\tableofcontents

\input{sections/introduction}

\input{sections/preliminaries}

\input{sections/kkthn_framework}

\input{sections/applications}

\input{sections/conclusions}

\section*{Acknowledgment} 
The authors gratefully acknowledge partial funding support from Texas A\&M University.

\section*{Declaration of Generative AI and AI-Assisted Technologies}
The authors used generative artificial intelligence (AI) tools to  assist with generating initial code, language editing, grammar, readability, and clarity. Generative AI tools were not used to generate research data, perform scientific analyses, develop conclusions, or make scientific claims. All AI-assisted contents were reviewed, tested and modified by the authors, who take full responsibility for the scientific content and integrity of the manuscript.

\bibliographystyle{ieeetr} 
\begingroup
\setstretch{1.0}
\bibliography{References}
\endgroup

\newpage

\appendix
\addtocontents{toc}{\protect\setcounter{tocdepth}{1}}
\input{appendix/appendix_metrics_benchmarks}
\input{appendix/appendix_projection_solve}
\input{appendix/appendix_fb_reduced}
\input{appendix/appendix_control_analysis}
\input{appendix/appendix_adam_projection}
\end{document}

%% file: packages.tex
\usepackage[T1]{fontenc}

\usepackage{url}
\usepackage{xcolor}
\usepackage{color,soul}
\usepackage{graphicx}
\usepackage{footnote}
\usepackage{silence}
\usepackage{microtype}
\usepackage{textcomp}

\usepackage{amsmath}
\usepackage{amssymb}
\usepackage{amsfonts}
\usepackage{amsthm}
\usepackage{mathtools}
\usepackage{nccmath}
\usepackage{nicefrac}
\usepackage{gensymb}
\usepackage{eucal}
\usepackage{upgreek}

\usepackage{booktabs}
\usepackage{multirow}
\usepackage{adjustbox}
\usepackage{ltablex}
\usepackage{makecell}

\usepackage{float}
\usepackage{caption}
\usepackage{subcaption}
\usepackage{rotating}
\usepackage{wrapfig}
\usepackage{placeins}
\usepackage{lscape}

\usepackage{tcolorbox}
\usepackage{mdframed}
\usepackage{comment}
\usepackage{tikz}
\usetikzlibrary{arrows.meta, positioning}
\usetikzlibrary{arrows.meta,shapes.geometric}

\usepackage[ruled,vlined]{algorithm2e}

\SetAlCapSkip{0.5em}
\SetAlgoInsideSkip{smallskip}
\SetAlgoHangIndent{0pt}

\SetKwInput{Problem}{Problem}
\SetKwInput{Parameter}{Parameter}
\SetKw{Initialize}{initialize}
\SetKw{Compute}{compute}
\SetKwInput{Extract}{Extract}

\usepackage{enumitem}
\usepackage{nomencl}
\usepackage{fontawesome5}
\usepackage{setspace}
\usepackage{fancyhdr}
\usepackage{titlesec}
\usepackage{newtxmath}

\usepackage{hyperref}

\usepackage{listings}

\definecolor{codebackground}{RGB}{255,255,255}
\definecolor{codeframe}{RGB}{0,0,0}
\definecolor{codecomment}{RGB}{150,170,150}
\definecolor{codekeyword}{RGB}{0,50,200}
\definecolor{codestring}{RGB}{160,40,40}
\definecolor{codenumber}{RGB}{100,100,100}

\lstdefinestyle{manuscriptcode}{
    backgroundcolor=\color{codebackground},
    commentstyle=\color{codecomment},
    keywordstyle=\color{codekeyword}\bfseries,
    numberstyle=\scriptsize\color{codenumber},
    stringstyle=\color{codestring},
    basicstyle=\ttfamily\small,
    breakatwhitespace=false,
    breaklines=true,
    breakautoindent=true,
    breakindent=0pt,
    columns=flexible,
    keepspaces=true,
    captionpos=b,
    numbers=left,
    numbersep=8pt,
    showspaces=false,
    showstringspaces=false,
    showtabs=false,
    tabsize=4,
    frame=single,
    rulecolor=\color{codeframe},
    framerule=0.75pt,
    framesep=5pt,
    xleftmargin=2em,
    xrightmargin=0pt,
    framexleftmargin=0pt,
    framexrightmargin=0pt,
    aboveskip=8pt,
    belowskip=8pt
}

%% file: sections/introduction.tex
\section{Introduction} \label{intro}
Mathematical models provide representations of real-world problems that help describe systems behavior and enable the analysis and identification of suitable solutions. 
However, constructing accurate mathematical models can be difficult for complex systems, particularly when the underlying phenomena and mechanisms are only partially known or computationally expensive to represent. In such cases, data-driven learning offers an attractive alternative, while existing domain knowledge can still provide valuable structure and physical consistency.
Purely data-driven machine learning such as multilayer perceptrons (MLPs) and deep neural networks (DNNs) can provide fast approximations, but its performance is often limited by scarce data, poor extrapolation, and the lack of guarantees that predictions remain physically meaningful. Incorporating domain knowledge, such as governing equations, conservation laws or other prior information, can reduce the effective search space and improve data efficiency, robustness, and generalization \cite{von2021informed,karniadakis2021physics}.

Scientific machine learning (SciML) has increasingly explored the integration of physical constraints into neural-networks (NNs). NNs are effective function approximators and have been widely used for surrogate modeling, physics-informed learning \cite{raissi2019physics}, inverse design, operator learning, and optimization. However, standard neural-network training is fundamentally unconstrained, which makes the direct enforcement of physical and operational constraints challenging. Penalty-based approaches address this by augmenting the loss function with weighted constraint violations, but fixed penalty weights generally do not guarantee exact feasibility or satisfaction of the stationarity conditions of the corresponding constrained problem.
This limitation has motivated the development of physics-constrained machine learning (PCML) methods. Different approaches have been proposed in literature to incorporate physics in machine learning models to some extent. These can be broadly categorized into two classes: (a) soft constrained neural networks and  (b) hard constrained neural networks.\\

Soft constrained neural networks, such as physics-informed neural networks (PINNs) \cite{karniadakis2021physics}, incorporate physical constraints by penalizing physics violations through the loss function during training. Due to the simplicity and ability to incorporate differential constraints, this method and its variants have been successfully used to model fluid flow \cite{cai2021physics, mao2020physics, abueidda2025deepokan}, biological systems \cite{yazdani2020systems, ahmadi2024ai}, heat transfer \cite{cai2021physics}, and uncertainty quantification \cite{zhu2019physics}, among others, where first-principles are available. This involves balancing between the data-fidelity and the first-principle violation through weights:

\begin{ceqn}
    \begin{equation} \label{eq:pinn_loss_fn}
        \min_{\boldsymbol{\uptheta}} \ \frac{1}{N}\sum_{i=1}^N \mathcal{L}\left(F_\textrm{NN}\left(\boldsymbol{\uptheta},\boldsymbol{x}\right)\right) + \lambda_R R + \lambda_p L_\textrm{physics}
    \end{equation}
\end{ceqn}

where, $L_\textrm{physics}$ is the physics loss to penalize the violation of the known first-principles or physical constraints, and $\lambda_p$ is the corresponding penalty weight. Soft constrained methods minimize both the objective function (data loss) and the constraint violation (physics loss) simultaneously as an unconstrained function with pre-defined multipliers for each loss term. Thus it diverges from the concept of minimizing the Lagrangian where the multipliers are also variables of the minimization problem. To this point, soft constrained neural networks never guarantee to satisfy the physical constraints. This often leads to unreliable prediction or inference.

Unlike soft constrained approaches, hard constrained neural networks aim to enforce constraints strictly by design. As modeling physical systems sometime require exact satisfaction of algebraic constraints that are highly nonlinear in nature, hard constrained methods have gained great attention in recent years in the scientific community. PCML combines the strengths of computationally efficient prediction capability of data driven neural networks with guaranteed satisfaction of physical laws and high accuracy through projection, thereby providing both accurate and efficient hybrid modeling capabilities \cite{iftakher2025physics}. These methods aim to overcome the limitations of MLPs and PINNs. For example, systems having safety critical constraints are usually inequality constraints that are not well suited for PINNs. The methodological approaches for hard constrained neural networks can be classified into two general categories: (i) predict-and-complete and (ii) projection-based methods \cite{mukherjee2026physics}. The predict-and-complete approaches first predict a subset of variables and completes the rest from the equality constraints \cite{beucler2021enforcing, zamzam2020learning}. An advance to this approach involves completion through explicit or implicit set of equations and then correction through internal gradient descent procedure to enforce inequalities \cite{donti2021dc3}. Projection-based methods exploits optimization theory to restore feasibility. The strategy involves first predicting unconstrained variables and then projecting them to a constrained manifold. Advances in this method include strict satisfaction of linear equalities \cite{chen2024physics}, linear inequalities \cite{flores2025enforcing, min2024hardnet}, convex constraints \cite{grontas2025pinet}, nonlinear equalities and inequalities \cite{iftakher2025physics, lastrucci2025enforce, roy2026nlpopt, goertzen2026hardnet++, mohammadi2026pl}. In this tutorial we focus on the projection-based method to enforce general nonlinear constraints.

There are several key advantages of projection based hard-constrained neural networks. First projection-based hard-constrained methods separate feasibility restoration from loss minimization. The projection layer enforces the constraints, while the neural-network backbone is trained primarily to minimize the loss, thereby reducing the burden on gradient-based learning. Second, the framework has a nested optimization structure: the inner problem projects the network output onto the feasible set, while the outer problem trains the network to minimize the fitting objective. This structure suggests potential connections to bilevel optimization that warrant further investigation. Third, the approach can extend to parametric optimization, where the active constraint set varies with the problem parameters. Through training across parameter realizations, the network can implicitly learn these active-set transitions, potentially providing useful approximations to the critical-region structure studied in multiparametric programming \cite{pistikopoulos2020multi}. With this realization, one can attempt to use hard-constrained neural networks to learn parametric solutions of optimization problems \cite{roy2026nlpopt}, where training encourages the neural network to generate predictions close to both feasible and optimal solution manifold. These predictions can therefore serve as high-quality warm starts for conventional optimization solvers, potentially reducing the number of iterations required for convergence. Hard-constrained learning also provides a promising approach for repeatedly solving parametric optimization problems at low inference cost, with applications in areas such as process control, optimal power flow, and trajectory optimization where rapid online decisions are required.

In this article, we present a tutorial of PCML with efficient and modular implementation of KKT-HardNet \cite{iftakher2025physics}, a neural-network-based PCML framework that enforces constraint feasibility through projection. We provide a step by step tutorial to allow users on how to use the KKT-HardNet code and the framework for various process system engineering applications. Among the novel contributions, we extend the original framework with multiple projection strategies, including the reduced-order Fischer--Burmeister formulation and a primal--dual interior-point method. To enable efficient end-to-end training, gradients through the converged projection are computed using implicit differentiation instead of autodifferentiation through the full sequence of solver iterations. We further provide an open-source, modular implementation with support for sparse computations to exploit special structure in large-scale problems. Finally, for deployment scenarios dominated by single-instance predictions, we support ahead-of-time (AOT) compilation to reduce runtime overhead and enable low-latency inference.

The tutorial is organised as follows: Section \ref{sec:kkthn} provides a foundational description of KKT-HardNet and its components along with implementation details. With the basic idea of the PCML framework, we then demonstrate how it can be used for hybrid modeling of complex system \ref{sec:hybrid-modeling}, real-time and online optimization \ref{sec:cstr-rto}, model predictive control \ref{sec:cstr-lmpc}, and PCML embedded greybox optimization \ref{sec:syngas}. Across these applications, KKT-HardNet is evaluated in terms of predictive accuracy, constraint satisfaction, and computational performance, with comparisons against existing state-of-the-art PCML approaches.

%% file: sections/preliminaries.tex
\section{\texttt{KKT-HardNet}: A Physics Constrained Machine Learning Framework}\label{sec:kkthn}
\begin{figure}
    \centering
    \includegraphics[width=0.9\linewidth]{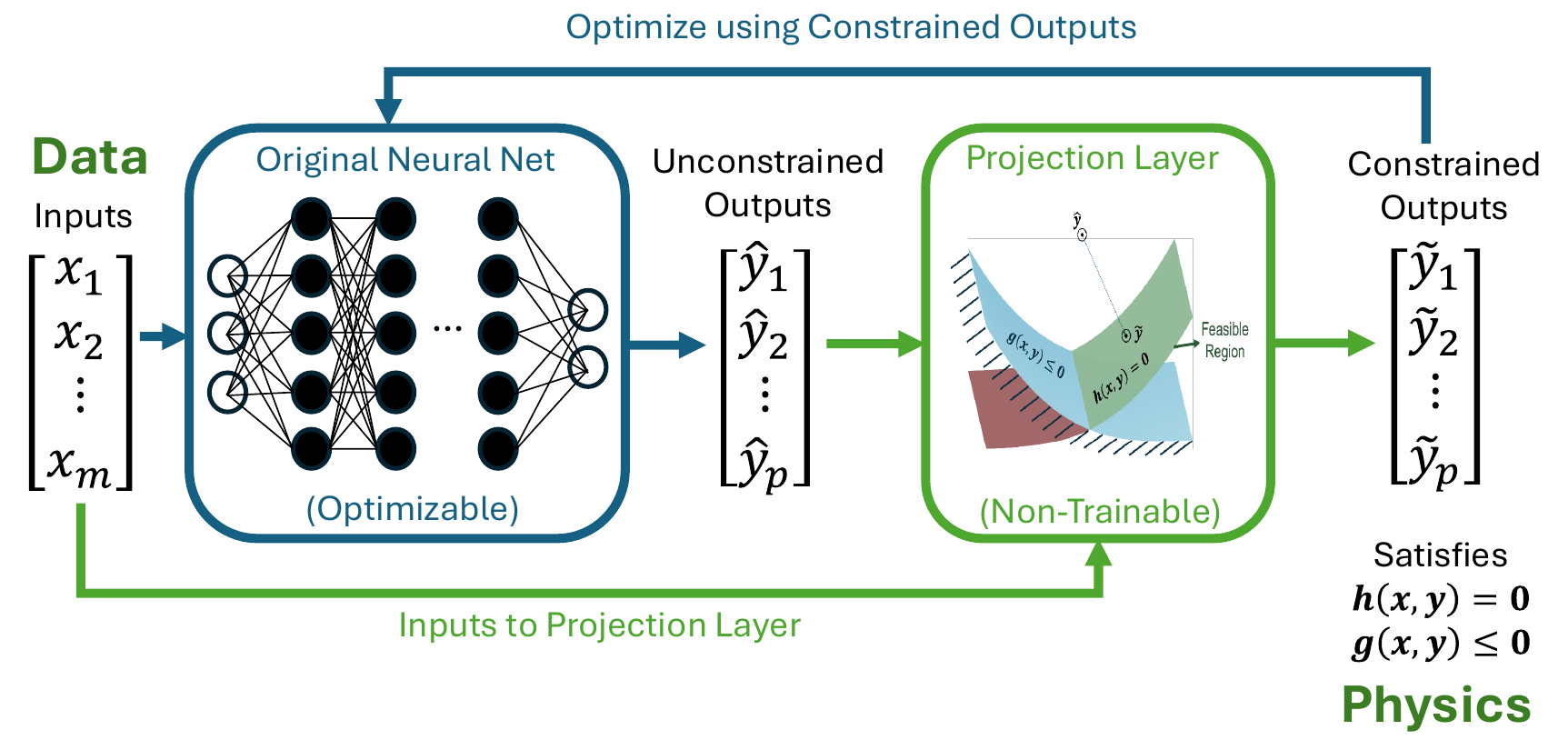}
    \caption{KKT-HardNet Architecture: The architecture takes parameters as input and through a backbone network predicts the raw output. The projection layer projects the raw output on the constrained surface which creates a feasible output. The model is trained with respective loss function on the constrained output.}
    \label{fig:architecture-diagram}
\end{figure}

In this section we describe a projection-based PCML framework, KKT-HardNet along with brief overview of different components. The framework is provided in Figure \ref{fig:architecture-diagram}. Given an unconstrained neural-network prediction, the projection layer solves an euclidian projection problem to obtain a nearby feasible point. This architecture can retain prediction accuracy while achieving constraint violations close to machine precision. We begin with exploring neural networks and their types followed by a brief introduction to projection layer and the training algorithm. Though machine learning covers a broad variety of tools, we limit our discussion in the space of neural networks.
Here, we briefly describe the different components of KKT-HardNet: a NN backbone, projection layer and the loss function. Note that, KKT-HardNet is a general framework where different architectures of neural network and projection algorithms can be integrated.

\subsection{Neural Network Backbone}
\begin{figure}[!htbp]
    \centering

    \begin{subfigure}[t]{0.50\textwidth}
        \centering
        \includegraphics[width=\linewidth]{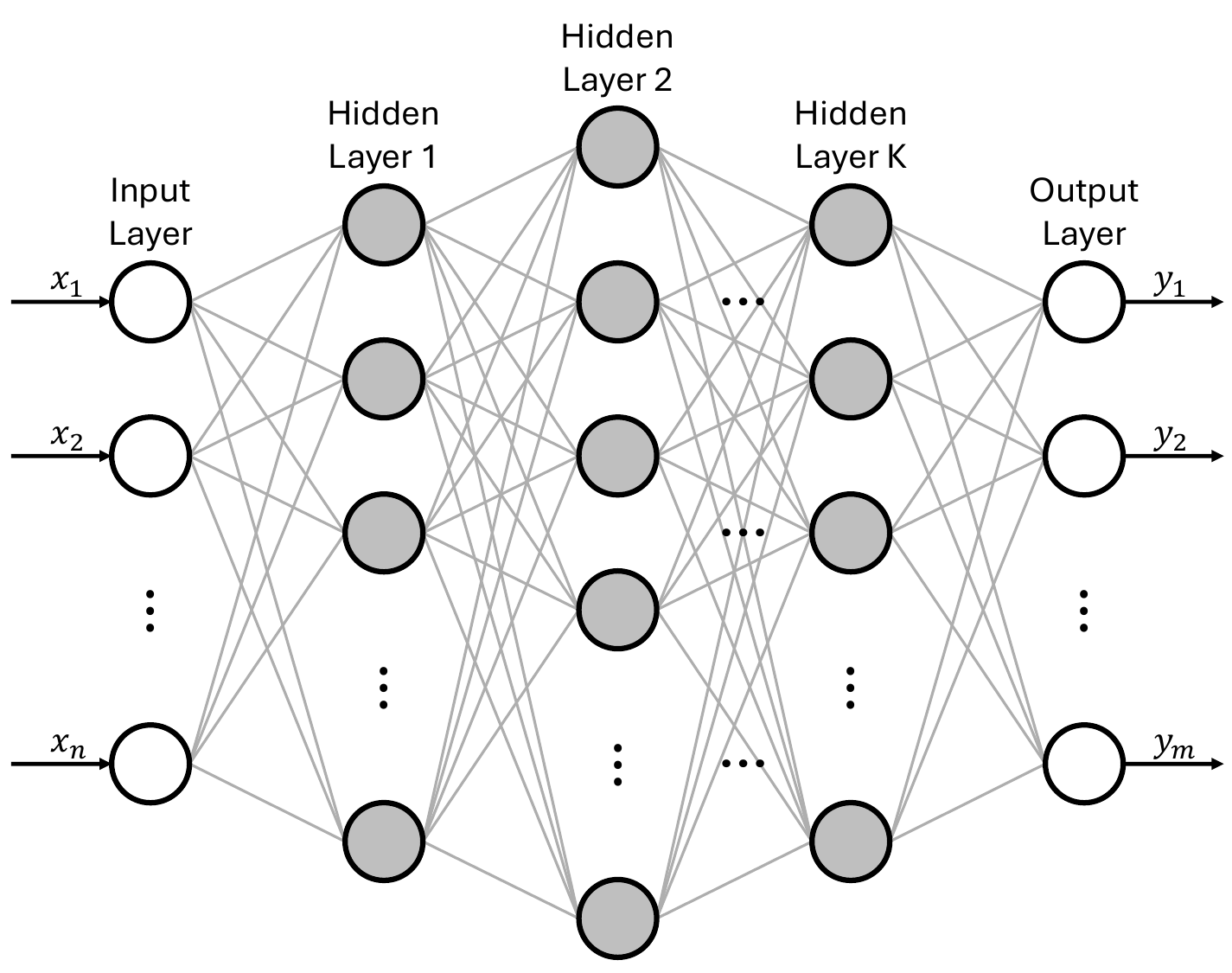}
        \caption{Multilayer perceptron}
        \label{fig:MLP-structure}
    \end{subfigure}
    \hfill
    \begin{subfigure}[t]{0.40\textwidth}
        \centering
        \includegraphics[width=\linewidth]{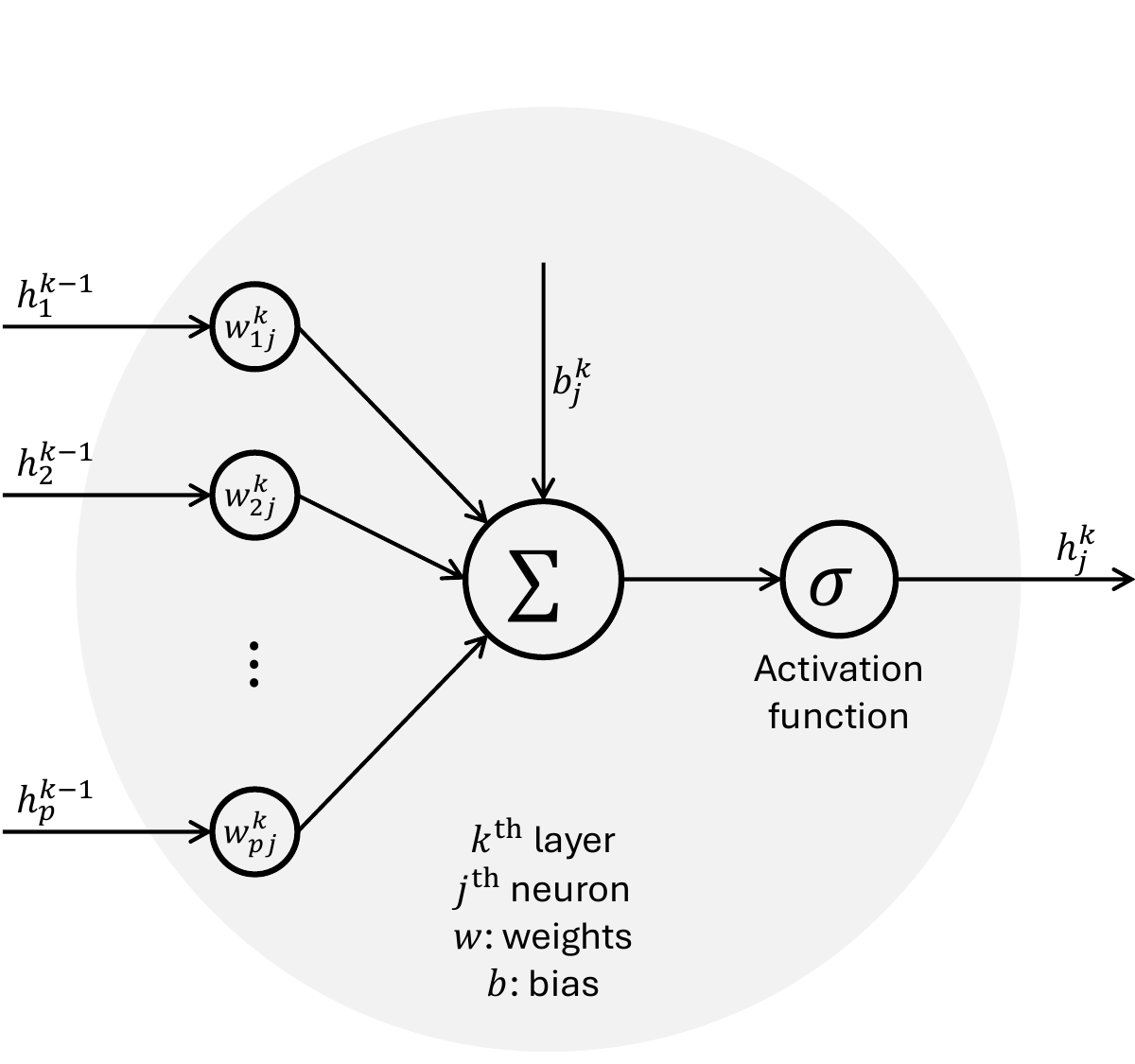}
        \caption{Neuron}
        \label{fig:perceptron}
    \end{subfigure}

    \caption{Schematic of a neural network.}
    \label{fig:NN_neuron}
\end{figure}

Neural networks (NNs) are a model structure to capture complex and nonlinear systems, originally inspired from mimicking how brains process information \cite{gurney2018introduction}. Despite being originated from neuroscience, NNs are now used widely due to their impressive capability of universally approximating any arbitrary function \cite{shah2025hybrid} (universal approximation theorem, see \cite{cybenko1989approximation} for details). NNs are perceived as networks consist of artificial neurons (known as perceptrons) arranged in layers \cite{haykin2009neural}. The simplest structured NNs are MLPs shown in Figure \ref{fig:MLP-structure}. Mathematically, neurons are made with a composition of linear and nonlinear functions, making it suitable for approximating general nonlinear functions. The layers in NNs are arranged sequentially and the output of each layer goes to the next layer as input. The first and last layers are input and output layers and the intermediate layers are hidden layers. The information given to the input layer are the input features ($\boldsymbol{x}\in\mathbb{R}^n$) and the information coming out of the output layer are the predicted outputs ($\boldsymbol{\hat{y}}\in\mathbb{R}^m$), where ``hat" represents unconstrained output. Let $h_j^k$ be the output of $j^\textrm{th}$ neuron in $k^\textrm{th}$ hidden layer (shown in Figure \ref{fig:perceptron}), which is computed as 

\begin{ceqn}
    \begin{equation}
        h_j^k = \sigma_j^k \left(\sum_{i=1}^{S_{k-1}}w^k_i h_i^{k-1}+b_j^k\right),
    \end{equation}
\end{ceqn}

where $S_{k-1}$ is the number of neurons in the $(k-1)^\textrm{th}$ layer ($S_{k-1}=p$ in Figure \ref{fig:perceptron}), $h_i^{k-1}$ are the inputs coming from ${(k-1)}^{th}$ layer, $w^k_{ij}$ are the weights for the $i$, $j$ neuron connection, $b_j^k$ is the bias associated with the particular neuron, and $\sigma$ is the activation function. For a MLP with $K$ layers, the functional form of the NN, output is represented as follows:

\begin{ceqn}
    \begin{equation}
        \boldsymbol{\hat{y}} = F_\textrm{NN}\left(\boldsymbol{\uptheta},\boldsymbol{x}\right) =\begin{cases}
            \begin{aligned}
            \boldsymbol{y} &= \boldsymbol{W}^{K+1} \boldsymbol{h}^{K}+\boldsymbol{b}^{K+1},\\
            \boldsymbol{h}^k &= \boldsymbol{\sigma}^k(\boldsymbol{W}^k \boldsymbol{h}^{k-1}+\boldsymbol{b}^{k}), \forall k = 1,2,\dots,K,\\
            \boldsymbol{h}^0 &= \boldsymbol{x},
        \end{aligned}
        \end{cases}
    \end{equation}
\end{ceqn}

where $\boldsymbol{\uptheta}$ represents the NN parameters, i.e., the weights and biases over which the NN is trained with a loss function and $\boldsymbol{h}^k$, $\boldsymbol{\sigma}^k$ and $\boldsymbol{b}^{k}$ are the hidden layer outputs, the activation functions, and the biases of the neurons in $k^\textrm{th}$ layer respectively. The matrix $\boldsymbol{W}^k$ are the weights for the neuron connection between $k^\textrm{th}$ and $(k-1)^\textrm{th}$ layers. Typically, $K+1$ refers to the output layer where generally the use of activation function is avoided (except for special cases). Commonly used activation functions are sigmoid, rectified linear unit (ReLU), hyperbolic tangent (tanh), etc. (see \cite{rasamoelina2020review} for details). The choice of activation function depends on the modeling purpose and the nonlinearity in the data, among others.

\subsection{Projection Layer}
Before defining the projection layer, we recall that our goal is to train a neural network $\boldsymbol{y} = F_\textrm{NN} \left( \boldsymbol{\uptheta}, \boldsymbol{x} \right)$ while satisfying the general nonlinear constraints of the form $\boldsymbol{h}(\boldsymbol{x}, \boldsymbol{y})=\boldsymbol{0}$ and $\boldsymbol{g}(\boldsymbol{x}, \boldsymbol{y}) \leq \boldsymbol{0}$, where $\boldsymbol{x}\in\mathbb{R}^n$ is the neural network input, $\boldsymbol{y}\in\mathbb{R}^m$ is the neural network output, $\boldsymbol{h}: \mathbb{R}^{n+m}\mapsto\mathbb{R}^p$ and $\boldsymbol{g}: \mathbb{R}^{n+m}\mapsto\mathbb{R}^q$ are the set of equality and inequality constraint functions with $p$ and $q$ constraints respectively. We make the following assumptions:

\begin{enumerate}
    \item the number of equality constraints is less than or equal to the number of neural network output, i.e., $p\leq m$ to ensure sufficient degrees of freedom for learning, and
    \item the constraint manifold $\mathcal{S}_\textrm{feasible}=\{y\vert \boldsymbol{h}(\bar{\boldsymbol{x}}, \boldsymbol{y})=\boldsymbol{0}, \boldsymbol{g}(\bar{\boldsymbol{x}}, \boldsymbol{y}) \leq \boldsymbol{0}\}$ is nonempty for a given $\bar{\boldsymbol{x}}$ and holds linear independence constraint qualification (LICQ).
\end{enumerate}

\begin{figure}[!htbp]
    \centering
    \includegraphics[width=0.7\linewidth]{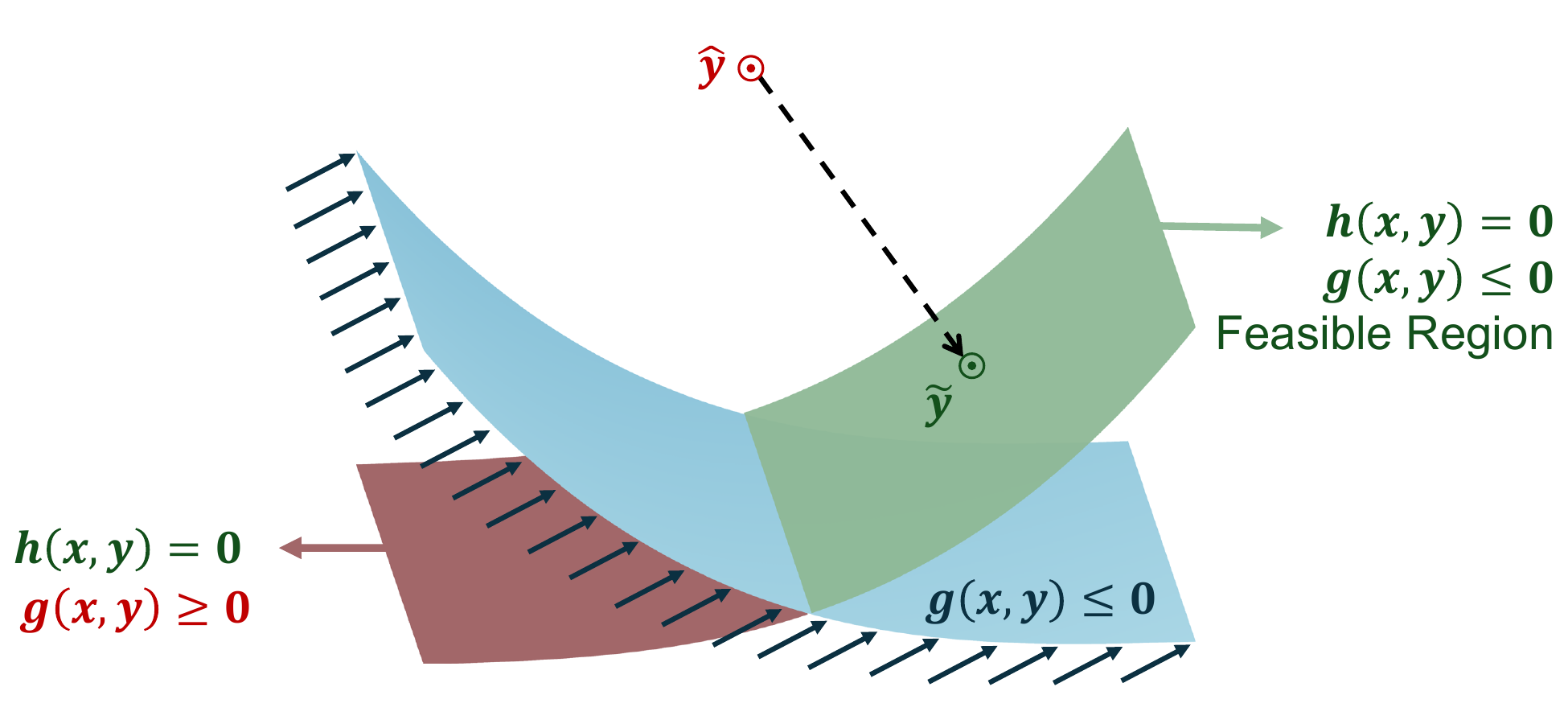}
    \caption{Graphical interpretation of the projection problem. The unconstrained prediction is pushed towards the feasible domain. While the global minima will represent the minimum distance solution, solving through KKT system guarantees only feasibility. \cite{iftakher2025physics}}
    \label{fig:projection}
\end{figure}

Let, $\hat{\boldsymbol{y}} = F_\textrm{NN}\left( \boldsymbol{\uptheta}, \bar{\boldsymbol{x}} \right)$ be the raw output from the neural network for a given $\bar{\boldsymbol{x}}$, the projection layer is defined as the following feasibility restoration step (commonly used in nonlinear optimization \cite{nocedal2006numerical}):

\begin{ceqn}
    \begin{equation} \label{eq:projection}
        \begin{aligned}
            \tilde{\boldsymbol{y}} = \mathcal{P}(\hat{\boldsymbol{y}},\boldsymbol{\bar{x}}) = \arg \min_{\boldsymbol{y}} \ &\frac{1}{2} \lVert \boldsymbol{y} - \hat{\boldsymbol{y}}\rVert^2\\
            \textrm{s.t.} \ & h_i(\bar{\boldsymbol{x}}, \boldsymbol{y}) = 0, && i=1,2,\dots,p\\
            & g_j(\bar{\boldsymbol{x}}, \boldsymbol{y}) \leq 0, && j=1,2,\dots,q,
        \end{aligned}
    \end{equation}
\end{ceqn}

where, $\mathcal{P}(\hat{\boldsymbol{y}},\boldsymbol{\bar{x}})$ is the implicit projection step. Mathematically, the solution to this optimization problem is the minimum distance feasible solution from the unconstrained and generally infeasible point $\hat{\boldsymbol{y}}$ (see Figure \ref{fig:projection}). $\tilde{y}$ is the ultimate output that is guaranteed to meet the physical constraints. 

While analytical solution to the projection problem in Eq. \ref{eq:projection} exists for only linear equality constraints, solving the problem for general nonlinear constraints is not trivial and require iterative steps. One way is considering the Karush-Kuhn-Tucker (KKT) conditions of the problem to find a feasible solution (in some cases the optimal solution) to guarantee constraint feasibility. See Appendix \ref{app:projection_solution} for brief details on solving the projection problem. While the efficiency of the forward projection problem is subject to the solution strategy used, efficiency of training relies on the backpropagation. Implicit function theorem can be utilized to make the training efficient with implicit backward gradient computation. With implicit differentiation, the projection algorithm may contain non-differentiable intermediate operations, provided that the converged solution admits a locally differentiable implicit representation. This substantially broadens the class of optimization algorithms that can be incorporated into the projection layer \cite{amos2017optnet}. 

\subsection{Loss Function and Backpropagation}
While the projection layer ensures that the final prediction, $\tilde{y}$, strictly satisfies the equality and inequality constraints, we now present an updated loss function for the NN, based on $\tilde{y}$, so that the NN also learns the physics. To that end, while various loss functions exist (see \cite{golder2026physics} for a broader discussion), training KKT-HardNet boils down to solving the structural risk minimization (SRM) problem of the following form:

\begin{ceqn}
\begin{equation}
\min_{\boldsymbol{\theta}}\quad
 \frac{1}{N}\sum_{i=1}^{N}
\mathcal{L}_{\theta}\!\left(\tilde{\boldsymbol{y}}_i\right)
+\lambda_R R(\boldsymbol{\theta}).
\label{eq:loss_fn}
\end{equation}
\end{ceqn}

where  $\tilde{\boldsymbol{y}}_i = F_{\mathrm{NN}}(\boldsymbol{\theta},\boldsymbol{x}_i)$ is the model prediction of the $i^{th}$ sample, $N$ is the number of training samples, $\mathcal{L}$ denotes the task-specific loss function, $R(\boldsymbol{\theta})$ is a regularization term, and $\lambda_R$ is the corresponding regularization weight. The choice of $\mathcal{L}$ depends on the learning task. For supervised surrogate modeling and parameter-estimation problems, the loss typically measures the discrepancy between the model prediction and the available target data. For optimization and control applications, the physical or economic objective of the underlying optimization problem can instead be used directly as the training loss. Training, therefore, consists of determining the network parameters $\boldsymbol{\theta}$ that minimize the selected task-specific objective while accounting for regularization. Representative loss functions and their corresponding applications considered in this work are summarized in Table~\ref{tab:loss-functions}. One important point to note that, depending on the choice of loss function in the PCML framework, one could deploy PCML for a wide range of process systems engineering capabilities (some are listed in Table \ref{tab:loss-functions}), including hybrid model development, learning solutions of parametric optimization problems, estimating parameters, etc, by steering the $\hat{y}$ towards the desired outcome. Finally projection corrects $\hat{y}$ to $\tilde{y}$.

\begin{table}[!htbp]
\centering
\caption{Representative loss functions, learning types, and corresponding applications in the KKT-HardNet framework.}
\label{tab:loss-functions}
\begin{tabular}{p{0.34\textwidth} p{0.16\textwidth} p{0.40\textwidth}}
\toprule
\textbf{Loss function} & \textbf{Type} & \textbf{Application Areas} \\
\midrule

$\displaystyle
\mathcal{L}_{\mathrm{data}}
=
\frac{1}{2N}\sum_{i=1}^{N}
\left\|
\tilde{\boldsymbol{y}}_i-\bar{\boldsymbol{y}}_i
\right\|_2^2
$
&
Supervised
&
Hybrid modeling, data-driven optimization, and parameter estimation
\\[10pt]

$\displaystyle
\mathcal{L}_{\mathrm{obj}}
=
\frac{1}{N}\sum_{i=1}^{N}
f\!\left(\boldsymbol{x}_i,\tilde{\boldsymbol{y}}_i\right)
$
&
Unsupervised
&
Process optimization, process control
\\

\bottomrule
\end{tabular}
\end{table}

Training NNs was popularized by the success of the error backpropagation algorithm \cite{rumelhart1986learning} which is a first-order algorithm to train deep neural networks. It computes the gradient of the total loss with respect to the NN parameters through chain rule and utilizes the gradient to update the parameters. While backpropagation is efficient in gradient computation, second order methods such as Gauss-Newton \cite{osborne1992fisher}, Levenberg-Marquardt \cite{levenberg1944method, marquardt1963algorithm} algorithms have also shown promise to train NNs.

%% file: sections/kkthn_framework.tex
%


\subsection{A Modular Python Package \texttt{(pip install kkt-hardnet)}}\label{kkthn}

\label{sec:kkthn-framework}

\begin{figure}[!htbp]
    \centering
    \includegraphics[width=\linewidth]{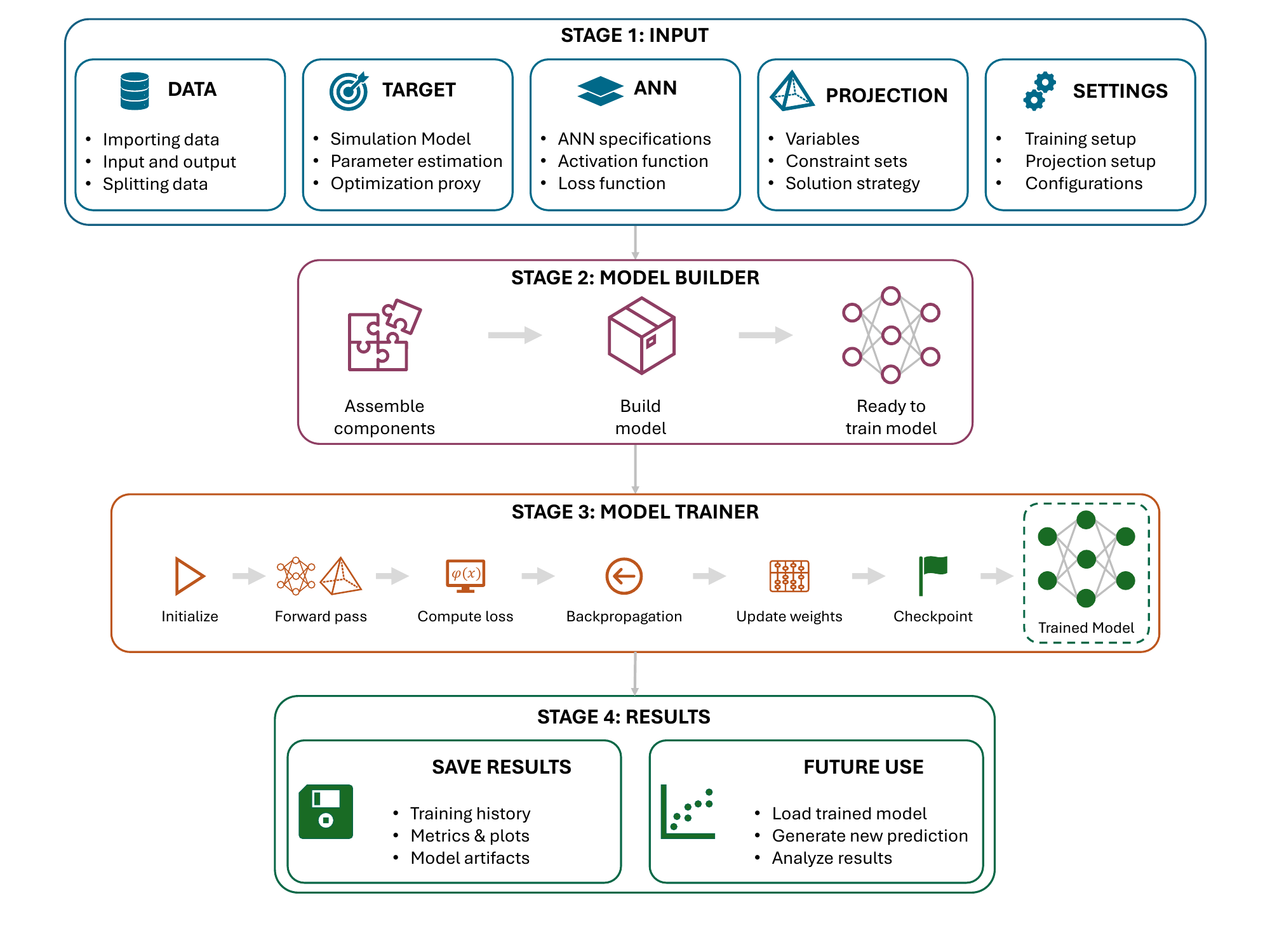}
    \caption{Modular organization of \texttt{KKT-HardNet}. Each row corresponds to a stage with different modules.}
    \label{fig:KKTHN-Package}
\end{figure}

\texttt{KKT-HardNet} is available as a Python package (\texttt{pip install kkt-hardnet}) and extends the original hard-coded proof-of-concept framework \cite{iftakher2025physics} into a modular implementation designed for practical use and extension. The package uses fixed-shape JAX computations so that the neural-network and projection operations can be compiled and executed together on a selected device \cite{jax2018github}. Figure ~\ref{fig:KKTHN-Package}, shows the four stages of the workflow: input, model building, model training, and results. The input stage specifies the data, constraints, neural-network architecture, and training/projection configurations; the model builder assembles these components; the training stage optimizes the neural-network backbone through the projected outputs while keeping the projection layer non-trainable; and the results stage saves the trained model, artifacts, and training history for future use. To support this workflow, \texttt{KKT-HardNet} provides a unified interface for defining neural-network backbones, differentiable constraints, projection methods, loss functions, and training procedures through six core modules: \texttt{problems}, \texttt{nn}, \texttt{losses}, \texttt{opt}, \texttt{train}, and \texttt{export}. Their responsibilities are summarized below:

The major modules of KKT-HardNet package are:
\begin{itemize}
    \item \texttt{problems}: This module includes robust ways of defining physics and operational constraints starting from symbolically representing the variables. Equations can be defined individually and as blocks with matrix form. We provide options for defining specific structured constraints such as linear or quadratic constraints. While simulation/surrogate modeling does not involve any optimization objective, for optimization proxy this module includes an option to define objective. Furthermore, variable bounds can be included separately as a special set of constraints.
    \item \texttt{nn}: This module utilizes \texttt{jax.nn} and includes common functions for NN libraries to define the NN backbone of the model. User has control over the architecture including selection of the activation function and the size of the NN backbone. Some commonly used activation functions are \texttt{sigmoid}, \texttt{tanh}, \texttt{relu}, etc.
    \item \texttt{losses}: This module deals with defining task specific loss function. For example, simulation/surrogate modeling, and parameter estimation involves data fitting for which mean squared error loss is defined with a consistency loss. For optimization, this creates a loss function that incorporates the objective function of the optimization problem.
    \item \texttt{opt}: This module includes different strategies for solving the projection problem. In our implementation, we have three methods: (i) explicit projection when only linear equality constraints are present, (ii) solving KKT system with Fishcer-Burmeister reformulation which gives a KKT-based feasible solution, and (iii) primal-dual interior point algorithm \footnote{We use the available open source code of the implementation to reproduce in high level python to make it suitable for batch operations and to include in training loop with implicit backward gradient.} \cite{wachter2006implementation}. While the explicit projection is a single step computation after unconstrained prediction, methods (ii) and (iii) involve iterative procedure to find a solution. Backward gradient computation is expensive for large scale problems through automatic differentiation. In out implementation, we leverage implicit function theorem to compute the implicit backward gradient for the projection layer (see \cite{amos2017optnet, grontas2025pinet, roy2026nlpopt} for details on implicit backward gradient). We also introduced a reduced method to solve KKT system with Fishcer-Burmeister reformulation which we describe in Appendix \ref{appendix:fishcer-burmeister-reduced}.
    \item \texttt{train}: The training stage takes the assembled pipeline and trains the NN backbone to minimize specified loss function. The `\texttt{train}' module includes the training and validation data along with the user defined training configurations to train and evaluate the model. This module also checks for the best found loss and saves that model for downstream use.
    \item \texttt{export}: This module provides the functionality required to serialize, compile, and load trained \texttt{KKT-HardNet} predictors for deployment. The implementation supports two artifact types: (i) a portable predictor represented using the JAX export format, and (ii) an environment-specific native predictor obtained by compiling the portable artifact for the current hardware and software environment. The portable loader deserializes the exported JAX computation and wraps it using a common predictor interface, allowing the model to be evaluated without reconstructing the original training code. A command line interface is additionally provided to compile a portable predictor into a native artifact directly on the target machine. The native artifact is target specific and requires a compatible JAX and \texttt{jaxlib} runtime. Therefore, it should not be interpreted as a completely standalone operating-system executable.
\end{itemize}
Codebox~\ref{lst:pseudocode} illustrates the corresponding problem-definition workflow. GPU support is available through \texttt{pip install "kkt-hardnet[cuda12]"}.

\begin{lstlisting}[
    style=manuscriptcode,
    language=Python,
    caption={Code Snippet for using KKT-HardNet.},
    label={lst:pseudocode}
]
% ======================================================
% Step 1: Import Necessary Libraries
% ======================================================
import os
os.environ["JAX_ENABLE_X64"] = "True"
from pathlib import Path
import numpy as np

import jax
jax.config.update("jax_enable_x64", True)
import jax.numpy as jnp

# Imports from KKT-HardNet
from kkthn import KKTHardNet, Loss, MLP, Projection

% ======================================================
% Step 2: Define Configurations
% ======================================================
SEED = 42
TRAIN = {# Define your Configuration here}
DATA_PATH = "your-data.csv"

% ======================================================================
% Step 3: Model Components: Backbone, Projection Layer and Loss Function
% ======================================================================
# Backbone
nn = MLP(hidden_sizes=(64, 64), activation="relu")

# Projection Layer
projection = Projection(name="my_model", method="fischer-burmeister-reduced")

x = projection.add_parameter(["x1", "x2", "x3"])
y = projection.add_variable(["y1", "y2", "y3"])
projection.constraints.add(
    y.y1 + y.y2 + y.y3 == 1,
    y.y1 >= 0, y.y1 <= 10,
    # Define other constraints)

# Loss Function
loss = Loss.mse()

% ======================================================================
% Step 4: Model Definition
% ======================================================================
model = KKTHardNet("surrogate", nn, loss, projection, train=TRAIN)
model.dataset(parameters=DATA_PATH, variables=DATA_PATH)

% ======================================================================
% Step 3: Model Training
% ======================================================================
result = model.model()

model.summary()
model.plot_history()
\end{lstlisting}

In the subsequent sections, we explore different applications of KKT-HardNet across different engineering fields. We include application codes to provide a self-contained tutorial and minimize reliance on external resources. Across all the case studies we present, a significant portion of the code remains same. These include library imports, backbone setup, training configurations, and solver configurations. See appendix \ref{app:benchmark} for the details on the performance metrics and the methods we consider to benchmark KKT-HardNet against.

%% file: sections/applications.tex
\section{PCML for Hybrid Modeling} \label{sec:hybrid-modeling}
Hybrid modeling combines data-driven learning with available first-principles knowledge to develop predictive models of complex systems. This is particularly useful for engineering processes where a complete mechanistic model may be difficult or computationally expensive to solve, while process data are available from experiments, simulations, or sensors. Although a neural network can learn the input--output relationship directly from these data, its predictions are not inherently guaranteed to satisfy known physical relations. KKT-HardNet provides a systematic way to incorporate such partial knowledge as hard constraints while retaining the flexibility of data-driven learning.

\begin{figure}[!htbp]
    \centering
    \includegraphics[width=\textwidth]{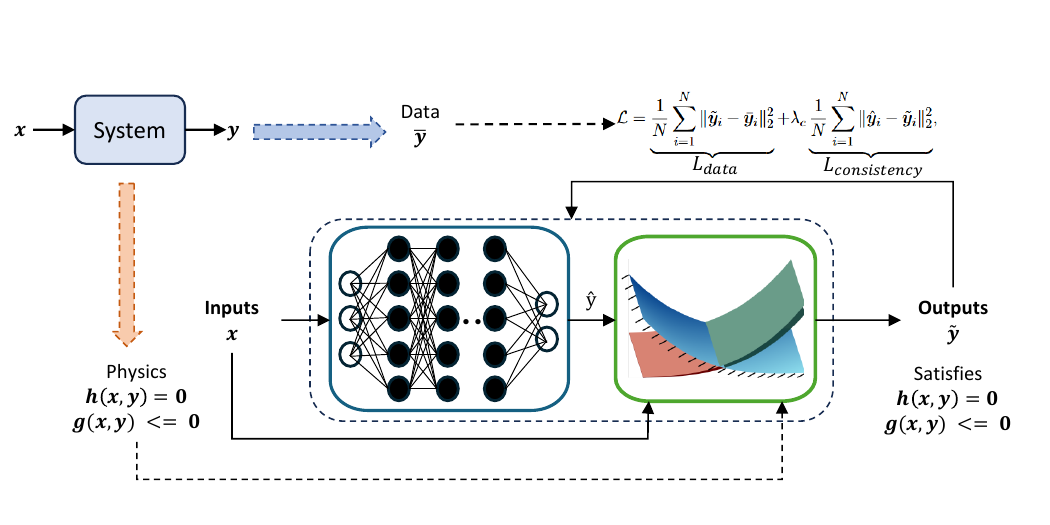}
    \caption{General KKT-HardNet representation for hybrid modeling, where available process data are combined with partial physical knowledge to learn a constrained input--output mapping.}
    \label{fig:reactor_condensor_diagram}
\end{figure}

In general, let $\boldsymbol{x}\in\mathbb{R}^{n_x}$ denote the observed process inputs and $\bar{\boldsymbol{y}}\in\mathbb{R}^{n_y}$ denote the corresponding measured or simulated outputs. A neural-network backbone learns the mapping $\tilde{\boldsymbol{y}}=F_{\mathrm{NN}}(\boldsymbol{\theta},\boldsymbol{x})$ from the available data, while known physical relations are represented by $h(\boldsymbol{x},\boldsymbol{y})=0$ and $g(\boldsymbol{x},\boldsymbol{y})\leq0$. KKT-HardNet projects the backbone prediction onto this constraint set, such that the final prediction $\boldsymbol{y}$ satisfies the incorporated physics. Importantly, the complete governing model is not required; only the available physical knowledge needs to be included in the projection layer. This allows the remaining relationships to be learned from data and makes the framework suitable for partially known systems. The general formulation of hybrid modeling with KKT-HardNet is illustrated in \ref{fig:reactor_condensor_diagram}.

\subsection{Hybrid Modeling of Chemical Processes} \label{app:comparison}

Here, we show the numerical and computational improvement of the framework compared to the base study reported by Iftakher et al. \cite{iftakher2025physics} (available at \href{https://github.com/SOULS-TAMU/kkt-hardnet/tree/old-workspace}{github.com/souls-tamu/kkt-hardnet/tree/old-workspace}). We compare two case studies: (i) Chemical process involving extractive distillation and (ii) Pooling problem involving nonlinear equality and inequality. For details of the systems we refer to \cite{iftakher2025physics}. Code for these case studies are available at the GitHub repository.

\subsubsection{A case study on chemical process simulation involving nonlinear thermodynamics}

\textbf{Key information:} 3 input variables, 9 output variables, 6 equality constraints, zero inequality constraints, includes nonlinearity.

\begin{figure}[htbp]
    \centering
    \begin{minipage}[t]{0.30\textwidth}
        \vspace{0pt}
        \centering
        \includegraphics[
            width=\linewidth,
            keepaspectratio
        ]{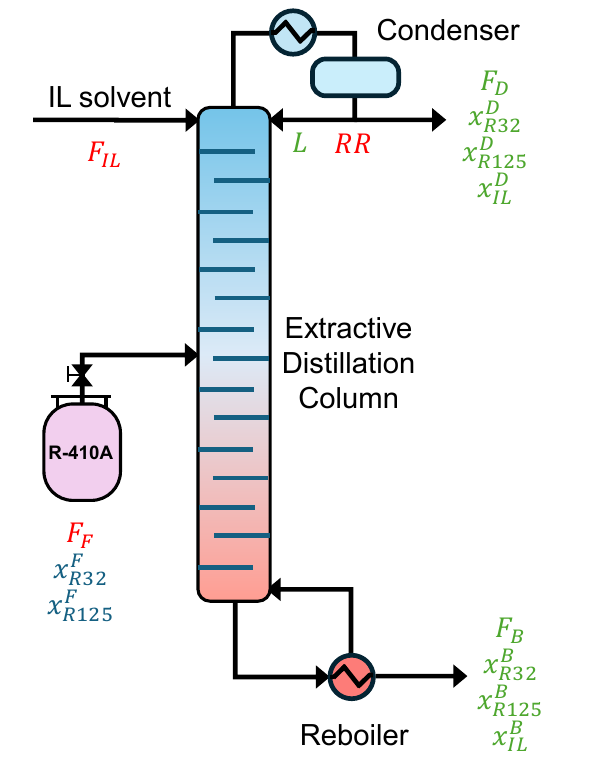}
        \captionof{figure}{Extractive distillation of R-410A using an Ionic liquid [EMIM][SCN].}
        \label{fig:ed_col}
    \end{minipage}
    \hfill
    \begin{minipage}[t]{0.68\textwidth}
        \vspace{0pt}
        \centering
        \captionof{table}{Performance comparison on the extractive distillation column case study}
        \label{tab:ed_col}

        \begin{tabular*}{\linewidth}{
            @{\extracolsep{\fill}}
            llllll
            @{}
        }
            \toprule
            & & &
            \multicolumn{3}{c}{KKT-HardNet (this tutorial)} \\
            \cmidrule(lr){3-5}
            Metric      & Base                                 & FB-N                  & FB-R                  & PDIP \\ \midrule
            T. MSE      & $1.05\times10^{-4}$                         & $1.10\times10^{-7}$   & $5.67\times10^{-8}$   & $3.27\times10^{-8}$\\
            V. MSE      & $1.08\times10^{-4}$                        & $3.16\times10^{-8}$   & $3.07\times10^{-8}$   & $1.72\times10^{-8}$\\
            T. Viol.    & $2.0\times10^{-8}$                         & $1.12\times10^{-7}$   & $6.60\times10^{-7}$   & $7.87\times10^{-13}$\\
            V. Viol.    & $2.0\times10^{-8}$                         & $5.53\times10^{-8}$   & $7.05\times10^{-7}$   & $9.71\times10^{-13}$\\
            Time (s)    & $12381.94$                                  & $122.92$               & $98.76$               & $108.36$\\
            \bottomrule
        \end{tabular*}
        \vspace{0.5em}
        \raggedright
        \footnotesize

        \textit{Abbreviations}\\
        FB-N: fischer-burmeister-newton\\
        FB-R: fischer-burmeister-reduced\\
        PDIP: primal-dual-interior-point\\
        T. MSE: Train MSE; V. MSE: Validation MSE\\
        T. Viol.: Train Violation; V. Viol.: Validation Violation\\
        Violation is measured as the mean absolute violation. Base studies from \cite{iftakher2025physics} were run on the same hardware for fair comparison.
    \end{minipage}
\end{figure}

\subsubsection{A case study on pooling problem involving nonlinear equality and inequality constraints}

\textbf{Key information:} 4 input variables, 5 output variables, 4 equality constraints, 2 inequality constraints, includes nonlinearity.

\begin{figure}[htbp]
\centering
    \begin{minipage}[t]{0.38\textwidth}
        \vspace{0pt}
        \centering
        \includegraphics[
            width=\linewidth,
            keepaspectratio
        ]{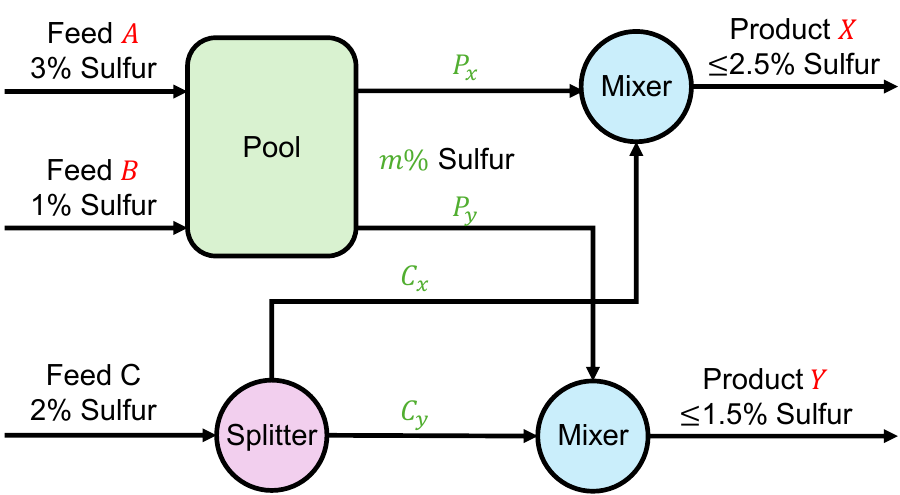}
        \captionof{figure}{Process flowsheet of the pooling problem. The product can be formed from three feeds.}
        \label{fig:example}
    \end{minipage}
    \hfill
    \begin{minipage}[t]{0.60\textwidth}
        \vspace{0pt}
        \centering
        \captionof{table}{Performance comparison on the pooling problem case study}
        \label{tab:example}

        \begin{tabular*}{\linewidth}{
            @{\extracolsep{\fill}}
            lllll
            @{}
        }
            \toprule
            & &
            \multicolumn{3}{c}{KKT-HardNet (this tutorial)} \\
            \cmidrule(lr){3-5}
            Metric      & Base                  & FB-N                  & FB-R                  & PDIP \\ \midrule
            T. MSE      & 49.65                 & 54.99                 & 55.05                 & 54.98 \\
            V. MSE      & 69.11                 & 45.43                 & 43.20                 & 43.49 \\
            T. Viol.    & $1.53\times10^{-5}$  & $1.51\times10^{-7}$   & $2.55\times10^{-7}$   & $4.99\times10^{-11}$ \\
            V. Viol.    & $1.53\times10^{-5}$  & $1.44\times10^{-7}$   & $3.25\times10^{-7}$   & $6.53\times10^{-11}$ \\
            Time (s)    & 8224.61               & 103.72                & 76.44                    & 68.07 \\
            \bottomrule
        \end{tabular*}
        \vspace{0.5em}
        \raggedright
        \footnotesize
        
    \end{minipage}
\end{figure}

\subsection{Hybrid Modeling of a Two-Phase Reactor-Condensor System}
\begin{figure}[!htbp]
    \centering
    \includegraphics[width=0.55\textwidth]{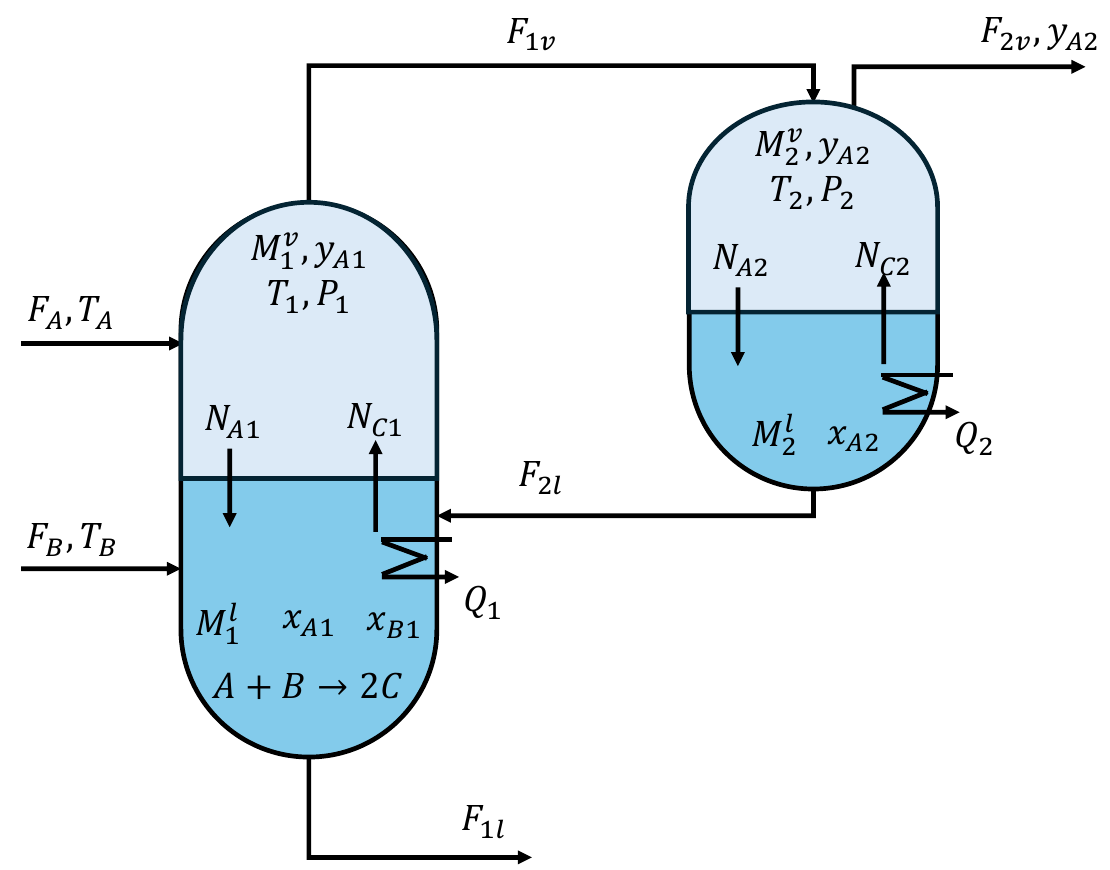}
    \caption{A two-phase Reactor-Condensor Network with recycle}
    \label{fig:reactor_condensor flowchart}
\end{figure}

To illustrate the hybrid modeling capability of a complex process simulation, we consider a two-phase reactor–condenser system with liquid recycle \cite{kumar1999control}, (illustrated in Figure \ref{fig:reactor_condensor flowchart}). The reactor pressure dynamics and interphase mass-transfer dynamics are assumed to be fast. Reactants $A$ and $B$ are fed to the reactor at molar flow rates $F_{A}$ and $F_{B}$ and temperatures $T_{A}$ and $T_{B}$, respectively, with $A$ entering in the vapor phase and $B$ in the liquid phase. Inside reactor, reactant $A$ transfers to the liquid phase at a rate $N_{A1}$, where the exothermic reaction $A+B\to 2C$ occurs.
\begin{ceqn}
    \begin{equation}
        R_{A}=k_{10}\exp\left(\frac{-E_{a}}{RT_{1}}\right)M_{1}^{l}\rho x_{A1}x_{B1},
    \end{equation}
\end{ceqn} 

\begin{table}[!htbp]
    \centering
    \caption{Parameter values and steady-state values for the reactor-condenser system}
    \begin{tabular}{ccc}
    \toprule
        Variable & Description & Nominal value \\
    \midrule
        $a$ & mass transfer area per unit liquid holdup volume $(m^{2}/m^{3})$ & $1000$ \\
        $k_{A}, k_{C}$ & overall mass transfer coefficients for A and C $(mol/(m^{2}\cdot s))$ & $20,\ 30$ \\
        $V_{1T}, V_{2T}$ & total volume of reactor and condenser $(m^{3})$ & $3.0,\ 3.0$ \\
        $\rho$ & liquid molar density $(kmol/m^{3})$ & $15.0$ \\
    \bottomrule
    \end{tabular}
    \label{tab:reactor_condensor_constants}
\end{table}


The training dataset was generated by solving a DAE model at steady state for a range of operating conditions. The problem property values are given in table \ref{tab:reactor_condensor_constants}. Six process inputs, $F_A$ and $F_B$, $T_A$ and $T_B$, $F_{2l}$, $F_{1l}$, were varied using Latin hypercube sampling to provide a stratified coverage of the prescribed operating domain. $F_A$, $F_B$, $F_{2l}$, and $F_{1l}$ were sampled within $\pm 15\%$ of their nominal values, while $T_A$ and $T_B$ were sampled over $305$--$325$ K and $290$--$310$ K, respectively. For each sampled input combination, the DAE was solved for steady state conditions using the IDA DAE solver from scikit-SUNDAE. Parameter continuation was employed by moving from the nominal operating point to each sampled condition through 12 intermediate steps, with the solution from each step used as the initial guess for the subsequent step. A solution was retained only when the IDA calculation converged, all predicted variables were finite, and the maximum absolute residual of the 19 DAE equations was less than $10^{-7}$. To keep the problem underdefined, we do not consider the steady state conditions as constraints in the projection layer. The following thermodynamic constraints are considered as equality constraints:
\begin{ceqn}
\begin{subequations}
\label{eq:reactor_condensor_equalities}
\begin{align}
0 &= N_{A1}-k_Aa(y_{A1}-y_{A1}^{*})\frac{M_1^l}{\rho},
& 0 &= N_{C1}-k_Ca\left(y_{C1}^{*}-(1-y_{A1})\right)\frac{M_1^l}{\rho}, \\
0 &= N_{A2}-k_Aa(y_{A2}^{*}-y_{A2}^{*})\frac{M_2^l}{\rho},
& 0 &= N_{C2}-k_Ca(1-y_{A2}-y_{C2}^{*})\frac{M_2^l}{\rho}, \\
0 &= P_1\left(V_{1T}-\frac{M_1^l}{\rho}\right)-M_1^vRT_1,
& 0 &= P_2\left(V_{2T}-\frac{M_2^l}{\rho}\right)-M_2^vRT_2, \\
0 &= P_1-P_2-\frac{1}{0.09}(F_{1v})^{7/4}. &&
\end{align}
\end{subequations}
\end{ceqn}

Only interphase mass transfer and pressure drop correlations are considered in the projection layer to keep the system underdefined. Additionally, the following inequality constraints are applied:
\begin{ceqn}
    \begin{equation}
        \begin{aligned}
            &M_{1}^{l} \in [5.0, 44.9] \quad M_{1}^{v} \in [0.01, 5.0] \\
            &M_{2}^{l} \in [0.1, 44.0] \quad M_{2}^{v} \in [0.3, 6.5] \\
            &x_{A1} \in [0.35, 0.70] \quad x_{B1} \in [0.18, 0.55] \quad x_{A1} + x_{B1} <= 1.0 \\
            &x_{A2} \in [0.55, 0.82] \quad y_{A1} \in [0.35, 0.56] \quad y_{A2} \in [0.22, 0.46] \\
            &T_{1} \in [315.0, 345.0] \quad T_{2} \in [280.0, 315.0] \\
            &P_{1} \in [45.0, 55.0] \quad P_{2} \in [45.0, 55.0] \\
            &F_{1v} \in [0.15, 0.27] \quad F_{2v} \in [0.10, 0.18] \\
            &N_{A1} \in [-0.02, 0.02] \quad N_{C1} \in [0.09, 0.13] \\
            &N_{A2} \in [0.03, 0.07] \quad N_{C2} \in [0.01, 0.03]
        \end{aligned}
        \label{eq:reactor_condensor_inequalities}
    \end{equation}
\end{ceqn}
to ensure an interior point as initialization from the backbone. We observe that without these inequality constraints the backbone produces a highly infeasible initialization which often diverges from the desired solution during projection. We use $1000$ different parameter conditions and split the dataset $85:15$ ratio for training and validation. The network backbone structure is given as $128$ hidden nodes and $2$ hidden layers with ReLU activation. We use ``Fischer-Burmeister-Reduced" formulation of the distance minimization problem with tolerance $10^{-9}$, maximum iterations $50$ and smoothing $10^{-9}$ along with other default configurations given in the code. We train the model for $3000$ epochs with a batch size of $50$, and learning rate $10^{-3}$. For the first $2000$ epochs we pretrain the backbone network for a suitable initialization to the projection. The training is supervised where we fit the data generated in the procedure mentioned before. We consider $7$ equality constraints \ref{eq:reactor_condensor_equalities} and $20$ inequalities \ref{eq:reactor_condensor_inequalities} in the projection layer. The code to train the model is given in codeblock \ref{lst:rc-pseudocode}. 

\begin{lstlisting}[
    style=manuscriptcode,
    language=Python,
    caption={Code for training the Reactor-Condensor model with KKT-HardNet.},
    label={lst:rc-pseudocode}
]
import os

from pathlib import Path
import jax
import jax.numpy as jnp
os.environ["JAX_ENABLE_X64"] = "True"
jax.config.update("jax_enable_x64", True)

from kkthn import KKTHardNet, Loss, MLP, Projection

SEED = 42

# Configurations
TRAIN = {"epochs": 700, "epoch_mlp": 600, "batch_size": 50, "learning_rate": 1e-3,
         "train_frac": 0.85, "seed": SEED, "dtype": "float64", "print_every": 100, 
         "minimum_projection_convergence": 0.95, "readiness_extra_epochs": 500, 
         "readiness_check_every": 10, "backend": "cpu", "output_dir": str(Path.cwd() / "runs")}

# Parameters and Variables
PARAMETER_NAMES = ["FA", "FB", "TA", "TB", "F2l", "F1l"]
VARIABLE_NAMES = ["M1l", "xA1", "xB1", "M1v", "yA1", "T1", "M2l", "xA2", "M2v", "yA2", "T2",
                    "NA1", "NC1", "NA2", "NC2", "P1", "P2", "F1v", "F2v"]
DATA_PATH = Path.cwd() / "reactor_condensor" / "reactor_condensor.csv"

# Backbone
nn = MLP(hidden_sizes=(128, 128), activation="tanh", predict_duals=True)

# Projection Layer
projection = Projection(name="reactor_condensor", method="fischer-burmeister-reduced",
    config={"smoothing": 1e-9, "tolerance": 1e-9, "max_iterations": 50})

x = projection.add_parameter(PARAMETER_NAMES)
y = projection.add_variable(VARIABLE_NAMES)

a, kA, kC, rho = 1000.0, 20.0, 30.0, 15.0
V1T, V2T, R_gas = 3.0, 3.0, 0.082057
MOL_TO_KMOL, PA_PER_ATM = 1.0e-3, 101325.0

PsA1 = projection.exp(25.1 - 3400.0 / (y.T1 + 20.0)) / PA_PER_ATM
PsC1 = projection.exp(27.3 - 4100.0 / (y.T1 + 70.0)) / PA_PER_ATM
PsA2 = projection.exp(25.1 - 3400.0 / (y.T2 + 20.0)) / PA_PER_ATM
PsC2 = projection.exp(27.3 - 4100.0 / (y.T2 + 70.0)) / PA_PER_ATM

yA1_star = PsA1 * y.xA1 / y.P1
yC1_star = PsC1 * (1.0 - y.xA1 - y.xB1) / y.P1
yA2_star = PsA2 * y.xA2 / y.P2
yC2_star = PsC2 * (1.0 - y.xA2) / y.P2

projection.constraints.add(
    y.NA1 - kA * a * (y.yA1 - yA1_star) * (y.M1l / rho) * MOL_TO_KMOL == 0,
    y.NC1 - kC * a * (yC1_star - (1.0 - y.yA1)) * (y.M1l / rho) * MOL_TO_KMOL == 0,
    y.NA2 - kA * a * (y.yA2 - yA2_star) * (y.M2l / rho) * MOL_TO_KMOL == 0,
    y.NC2 - kC * a * (1.0 - y.yA2 - yC2_star) * (y.M2l / rho) * MOL_TO_KMOL == 0,
    y.P1 * (V1T - y.M1l / rho) - y.M1v * R_gas * y.T1 == 0,
    y.P2 * (V2T - y.M2l / rho) - y.M2v * R_gas * y.T2 == 0,
    y.P1 - y.P2 - (1.0 / 0.09) * y.F1v ** (7.0 / 4.0) == 0
)

projection.constraints.add(
    y.M1l >= 5.0, y.M1l <= 44.9, y.M1v >= 0.01, y.M1v <= 5.0,
    y.M2l >= 0.1, y.M2l <= 44.0, y.M2v >= 0.3, y.M2v <= 6.5,
    y.xA1 >= 0.35, y.xA1 <= 0.70, y.xB1 >= 0.18, y.xB1 <= 0.55, y.xA1 + y.xB1 <= 1.0,
    y.xA2 >= 0.55, y.xA2 <= 0.82, y.yA1 >= 0.35, y.yA1 <= 0.56, y.yA2 >= 0.22, y.yA2 <= 0.46,
    y.T1 >= 315.0, y.T1 <= 345.0, y.T2 >= 280.0, y.T2 <= 315.0,
    y.P1 >= 45.0, y.P1 <= 55.0, y.P2 >= 45.0, y.P2 <= 55.0,
    y.F1v >= 0.15, y.F1v <= 0.27, y.F2v >= 0.10, y.F2v <= 0.18,
    y.NA1 >= -0.02, y.NA1 <= 0.02, y.NC1 >= 0.09, y.NC1 <= 0.13,
    y.NA2 >= 0.03, y.NA2 <= 0.07, y.NC2 >= 0.01, y.NC2 <= 0.03
)

# Loss function
loss = Loss.mse(primal_consistency_weight=1.0, equality_dual_consistency_weight=0.0,
                inequality_dual_consistency_weight=0.0)

# Model definition
model = KKTHardNet("surrogate", nn, loss, projection, train=TRAIN)

# Dataset
model.dataset(parameters=DATA_PATH, variables=DATA_PATH)

# Training
fb_reduced_result = model.model()
model.summary()
model.plot_history()

# Inference
model = KKTHardNet().load(Path("/path/to/metadata.json"))

x_new = jnp.array([[0.08981, 0.0533, 305.206, 308.4160856, 0.0678, 0.0108]])
y_pred = model.predict(x_new)
\end{lstlisting}

\begin{table}[!htbp]
\caption{Performance comparison for the reactor--condenser problem. We define the metrics in appendix \ref{app:benchmark}.}
\label{tab:reactor-condensor-performance}
\centering
\resizebox{\textwidth}{!}{%
\begin{tabular}{lrrrrrrr}
\toprule
 & MSE & Mean eq. & Max eq. & Mean ineq. & Max ineq. & Training Time (s) & Inference Time (ms/batch) \\
\midrule
MLP & $9.75 \times 10^{-2}$ & $0.29$ & $7.429$ & $2.22 \times 10^{-5}$ & $6.73 \times 10^{-2} $ & $39.78$ & $0.35$ \\
PINN & $1.31 \times 10^{-1}$ & $0.61$ & $5.29$ & $7.03 \times 10^{-3}$ & $9.80 \times 10^{-1}$ & $41.00$ & $0.31$  \\
ENFORCE & $7.21 \times 10^{0}$ & $7.36 \times 10^{-10}$ & $5.10 \times 10^{-9}$ & $0.00$  & $0.00$ & $16362.92$  & $208.39$ \\ 
KKT-HardNet & $3.11 \times 10^{-3}$ & $1.33 \times 10^{-7}$ & $8.59 \times 10^{-7}$ & $6.19 \times 10^{-8}$  & $9.45 \times 10^{-7}$ & $786.21$  & $57.86$ \\ \bottomrule
\end{tabular}
}
\end{table}
Table~\ref{tab:reactor-condensor-performance} compares the surrogate accuracy, constraint satisfaction, and computational performance of the different learning approaches. KKT-HardNet achieves the lowest prediction error, with an MSE of $3.11\times10^{-3}$, compared to other approached. At the same time, the projected predictions maintain small constraint residuals, with maximum equality and inequality violations of $8.59\times10^{-7}$ and $9.45\times10^{-7}$, respectively. Although ENFORCE attains tighter constraint residuals, it incurs substantially higher training and inference costs, requiring $16362.92$ s for training and $208.39$ ms per inference batch, compared to KKT-HardNet. KKT-HardNet takes only $786.21$ s and $57.86$ ms for training and inference, respectively. The unconstrained MLP and soft-constrained PINN are computationally cheaper, but exhibit considerably larger physical-constraint violations, demonstrating the trade-off between computational cost, predictive accuracy, and strict physical consistency.

\section{PCML for Real-Time Optimization}
\label{sec:cstr-rto}

Real-time optimization (RTO) determines economically optimal process decisions in response to changing process and economic conditions. Quantities such as energy prices, feed conditions, production targets, and equipment availability can vary with time, requiring the preferred operating point to change. RTO addresses this problem by repeatedly solving a constrained optimization problem while accounting for process physics and operating limits. In general, let $\boldsymbol{\theta}\in\mathbb{R}^{n_{\theta}}$ denote the time-varying process or economic parameters and $\boldsymbol{x}\in\mathbb{R}^{n_x}$ the decision variables. The RTO problem can be expressed as $\boldsymbol{x}^{*}(\boldsymbol{\theta})=\arg\min_{\boldsymbol{x}}f(\boldsymbol{\theta},\boldsymbol{x})$ subject to $h(\boldsymbol{\theta},\boldsymbol{x})=0$ and $g(\boldsymbol{\theta},\boldsymbol{x})\leq0$, where $f$ is the economic objective and $h$ and $g$ represent the process and operating constraints.

\begin{figure}[!htbp]
    \centering
    \includegraphics[width=\textwidth]{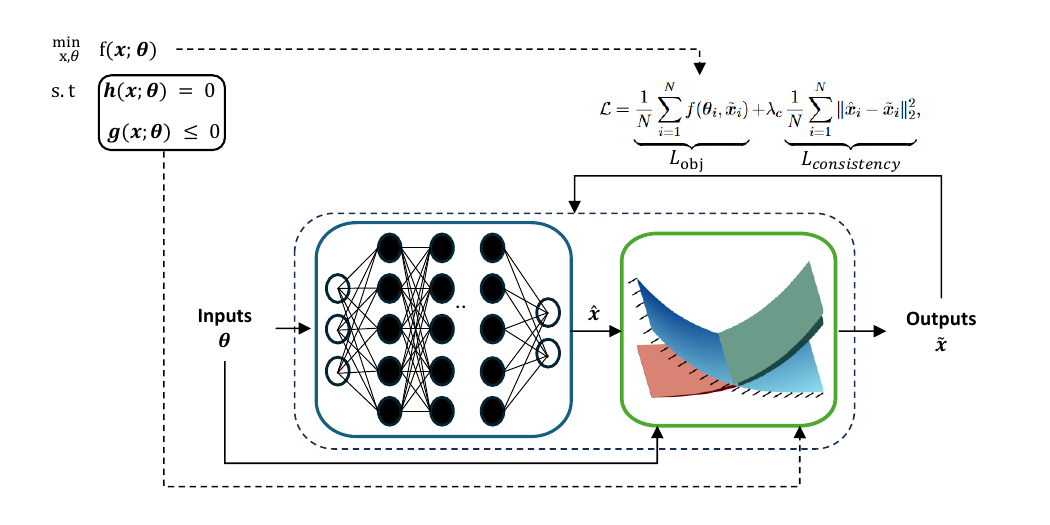}
    \caption{General KKT-HardNet formulation for real-time optimization, where the model learns the parametric solution map from the problem parameters $\boldsymbol{\theta}$ to a feasible near-optimal solution $\tilde{\boldsymbol{x}}$.}    \label{fig:rto_kkthn_diagram}
\end{figure}

Repeatedly solving such optimization problems can become computationally expensive, particularly for nonlinear systems and applications requiring frequent updates. KKT-HardNet provides an alternative by learning the parametric solution map $\boldsymbol{\theta}\mapsto\boldsymbol{x}^{*}$ offline. During training, the loss function drives the network toward economically favorable solutions, while the projection layer enforces the constraints. After training, changes in the process or economic parameters can be mapped directly to feasible near-optimal operating conditions through model inference, reducing the computational burden associated with repeatedly solving the original optimization problem online. The general RTO formulation with KKT-HardNet is illustrated in Figure \ref{fig:rto_kkthn_diagram}.

\subsection{A CSTR Optimization Case Study}
We consider a continuous stirred tank reactor (CSTR) with the reversible exothermic reaction $A\leftrightarrow B$ \cite{zhang2019real}. The process states are the concentrations $C_A$ and $C_B$ and the reactor temperature $T$, while $Q$ denotes the heat supplied through the reactor jacket. The nonlinear mass and energy balances are as follows:

\begin{ceqn}
\begin{equation}
\begin{aligned}
\frac{dC_A}{dt}
&=\frac{1}{\tau}(C_{A_0}-C_A)
-k_Ae^{-E_A/(RT)}C_A
+k_Be^{-E_B/(RT)}C_B,\\
\frac{dC_B}{dt}
&=-\frac{1}{\tau}C_B
+k_Ae^{-E_A/(RT)}C_A
-k_Be^{-E_B/(RT)}C_B,\\
\frac{dT}{dt}
&=-\frac{\Delta H}{\rho C_P}
\left(k_Ae^{-E_A/(RT)}C_A-k_Be^{-E_B/(RT)}C_B\right)
+\frac{1}{\tau}(T_0-T)
+\frac{Q}{\rho C_PV}.
\end{aligned}
\label{eq:cstr_dynamics}
\end{equation}
\end{ceqn}

Here, $T_0$ and $C_{A_0}$ denote the feed temperature and concentration; $k_A$ and $k_B$ are the pre-exponential factors; $E_A$ and $E_B$ are the activation energies; $\tau$ is the residence time; and $\Delta H$, $\rho$, $C_P$, and $V$ denote the reaction enthalpy, liquid density, heat capacity, and reactor volume, respectively. The model parameters are summarized in Table~\ref{tab:cstr_constants}.

\begin{table}[!htbp]
    \centering
    \caption{Parameter values and steady-state values for the CSTR case study.}
    \label{tab:cstr_constants}
    \begin{tabular}{cc}
    \toprule
        $T_0 = 400$ K & $\tau = 60$ s \\
        $k_A = 5000$ /s & $k_B = 10^{6}$ /s \\
        $E_A = 1\times10^{4}$ cal/mol & $E_B = 1.5\times10^{4}$ cal/mol \\
        $R = 1.987$ cal/(mol K) & $\Delta H = -5000$ cal/mol \\
        $\rho = 1$ kg/L & $C_P = 1000$ cal/(kg K) \\
        $C_{A_0} = 1$ mol/L & $V = 100$ L \\
        $C_{A_s} = 426.743$ K & $Q_s = 40386$ cal/s \\
    \bottomrule
    \end{tabular}
\end{table}

Because the reaction is reversible, excessively low temperature results in slow conversion, whereas high temperature increasingly favors the reverse reaction. The economically preferred operating point, therefore, depends on both reactant conversion and the cost of heat. We define the steady-state economic objective as

\begin{ceqn}
\begin{equation}
J_{\mathrm{RTO}}(C_A,Q;p)
=
\frac{C_A}{C_{A_0}}+pQ,
\label{eq:rto-cost}
\end{equation}
\end{ceqn}

where $p$ is the time-varying heat price. The corresponding parametric constrained RTO problem is

\begin{ceqn}
\begin{equation}
\begin{aligned}
\min_{C_A,C_B,T,Q}\quad
& \frac{C_A}{C_{A_0}}+pQ\\
\mathrm{s.t.}\quad
&\frac{1}{\tau}(C_{A_0}-C_A)
-k_Ae^{-E_A/(RT)}C_A
+k_Be^{-E_B/(RT)}C_B=0,\\
&-\frac{1}{\tau}C_B
+k_Ae^{-E_A/(RT)}C_A
-k_Be^{-E_B/(RT)}C_B=0,\\
&-\frac{\Delta H}{\rho C_P}
\left(k_Ae^{-E_A/(RT)}C_A-k_Be^{-E_B/(RT)}C_B\right)
+\frac{1}{\tau}(T_0-T)
+\frac{Q}{\rho C_PV}=0,\\
&0\le C_A\le1,\qquad
0\le C_B\le1,\qquad
400\le T\le500,\qquad
0\le Q\le10^5.
\end{aligned}
\label{eq:rto-optimization}
\end{equation}
\end{ceqn}

KKT-HardNet is trained directly as an optimization proxy for Eq.~\ref{eq:rto-optimization}, with heat price $p$ as the problem parameter ($\theta$) and $(C_A,C_B,T,Q)$ as the optimization variables ($x$). We sample $2500$ heat prices uniformly over $p\in[10^{-7},2\times10^{-6}]$ and use an $80{:}20$ training--validation split. The backbone consists of three hidden layers with $32$ neurons per layer and ReLU activation. The Fischer--Burmeister-Reduced projection is used with tolerance $10^{-10}$, maximum $300$ iterations, and smoothing parameter $10^{-10}$. The model is trained for $2000$ epochs using a batch size of $50$, learning rate $3\times10^{-4}$, and primal consistency weight ($\lambda_{c}$) to be $0.05$. The loss function is unsupervised and no label data is required. The implementation is shown in Code~\ref{lst:rto-pseudocode}.

\begin{lstlisting}[
    style=manuscriptcode,
    language=Python,
    caption={Code for training the real-time optimization (RTO) model with KKT-HardNet.},
    label={lst:rto-pseudocode}
]
import os
os.environ["JAX_ENABLE_X64"] = "True"
from pathlib import Path
import jax
jax.config.update("jax_enable_x64", True)
import jax.numpy as jnp
from kkthn import KKTHardNet, Loss, MLP, Projection

# System Parameters
SEED = 50, T_FEED = 400.0, CA_FEED = 1.0, TAU = 60.0, KA = 5000.0, KB = 1.0e6, EA = 1.0e4, 
EB = 1.5e4, R = 1.987, DEL_H = -5000.0, RHO = 1.0, CP = 1000.0, V = 100.0

# Constructor blocks
def equality_constraints(projection, y):
    rA = KA * projection.exp(-EA/(R*y.T_s))*y.CA_s
    rB = KB * projection.exp(-EB/(R*y.T_s))*y.CB_s
    r = rA-rB
    return [(CA_FEED-y.CA_s)/TAU-r==0.0,
            -y.CB_s/TAU+r==0.0,
            (-DEL_H/(RHO*CP)) * r + (T_FEED-y.T_s) / TAU + y.Q_s/(RHO*CP*V) == 0.0]

def inequality_constraints(y):
    return [y.CA_s >= 0.0, y.CA_s <= 1.0, y.CB_s >= 0.0, y.CB_s <= 1.0,
            y.T_s >= 400.0, y.T_s <= 500.0, y.Q_s >= 0.0, y.Q_s <= 1.0e5]

def objective_function(x, y):
    return y.CA_s / CA_FEED + x.heat_price * y.Q_s

# Configuration
TRAIN = { "epochs": 2000, "batch_size": 50, "learning_rate": 3e-4, "train_frac": 0.8,
          "seed": SEED, "dtype": "float64", "print_every": 10,
          "standardize_inputs": True, "pretrain": False, "drop_last_batch": True,
          "backend": "cpu", "output_dir": str(Path.cwd() / "runs")}
DATA_PATH = Path.cwd() / "rto" / "heat_price.csv"

# Backbone
nn = MLP(hidden_sizes=(32, 32, 32), activation="relu", predict_duals=True, inequality_dual_floor=1e-8)

# Projection Layer
projection = Projection(name="CSTR_RTO", method="fischer-burmeister-reduced",
    config={"tolerance": 1e-10, "max_iterations": 300, "smoothing": 1e-10})

x = projection.add_parameter(["heat_price"])
y = projection.add_variable(["CA_s", "CB_s", "T_s", "Q_s"])
projection.constraints.add(equality_constraints(projection, y))
projection.constraints.add(inequality_constraints(y))
projection.objective = objective_function(x, y)

# Loss function    
loss = Loss.objective(primal_consistency_weight=0.05, equality_dual_consistency_weight=0.00, inequality_dual_consistency_weight=0.00)
    
# Model and Training
model = KKTHardNet("optimization", nn, loss, projection, train=TRAIN)
model.dataset(parameters=DATA_PATH)
fb_reduced_result = model.optimize()

# Inference
model = KKTHardNet().load(Path("/path/to/metadata.json"))
x_new = jnp.array([[1.0e-6]])
y_pred = model.predict(x_new)
\end{lstlisting}

\begin{table}[!htbp]
\caption{Performance comparison for the CSTR real-time optimization problem in terms of objective value, constraint satisfaction, training time, and inference time for a batch of $100$ instances.}
\label{tab:rto_model_comparison}
\centering
\resizebox{\textwidth}{!}{%
\begin{tabular}{lrrrrrrr}
\toprule
 & Objective & Mean eq. & Max eq. & Mean ineq. & Max ineq.
 & Training Time (s) & Inference Time (ms/batch) \\
\midrule
IPOPT
& $0.535$ & $3.43\times10^{-12}$ & $9.98\times10^{-11}$
& $0.00$ & $0.00$ & $--$ & $1.83^*$ \\
PINN
& $-0.487$ & $1.96\times10^{-2}$ & $4.06\times10^{-2}$
& $0.06$ & $0.514$ & $56.87$ & $0.086$ \\
DC3
& $0.560$ & $1.11\times10^{-16}$ & $5.71\times10^{-18}$
& $0.00$ & $0.00$ & $333.38$ & $2.13$ \\
KKT-HardNet
& $0.535$ & $4.04\times10^{-8}$ & $8.57\times10^{-8}$
& $0.00$ & $0.00$ & $3725.66$ & $1.65$ \\
\bottomrule
\end{tabular}
}
\footnotesize
*Solver time is reported as average solve time considering full parallelization across 100 instances
\end{table}

As shown in Table~\ref{tab:rto_model_comparison}, KKT-HardNet reproduces the reference optimizer objective value of $0.535$ while maintaining very small tolerance for equality residuals and zero inequality violation. The PINN obtains an apparently lower objective but violates both equality and inequality constraints, whereas DC3 remains feasible but does not find the optimal objective value. KKT-HardNet also achieves smaller inference time of $1.65$ ms for a batch of $100$ instances, compared to $1.83$ ms for the reference optimizer.

\begin{figure}[!htbp]
    \centering
    \begin{subfigure}[t]{0.49\textwidth}
        \centering
        \includegraphics[width=0.8\linewidth]{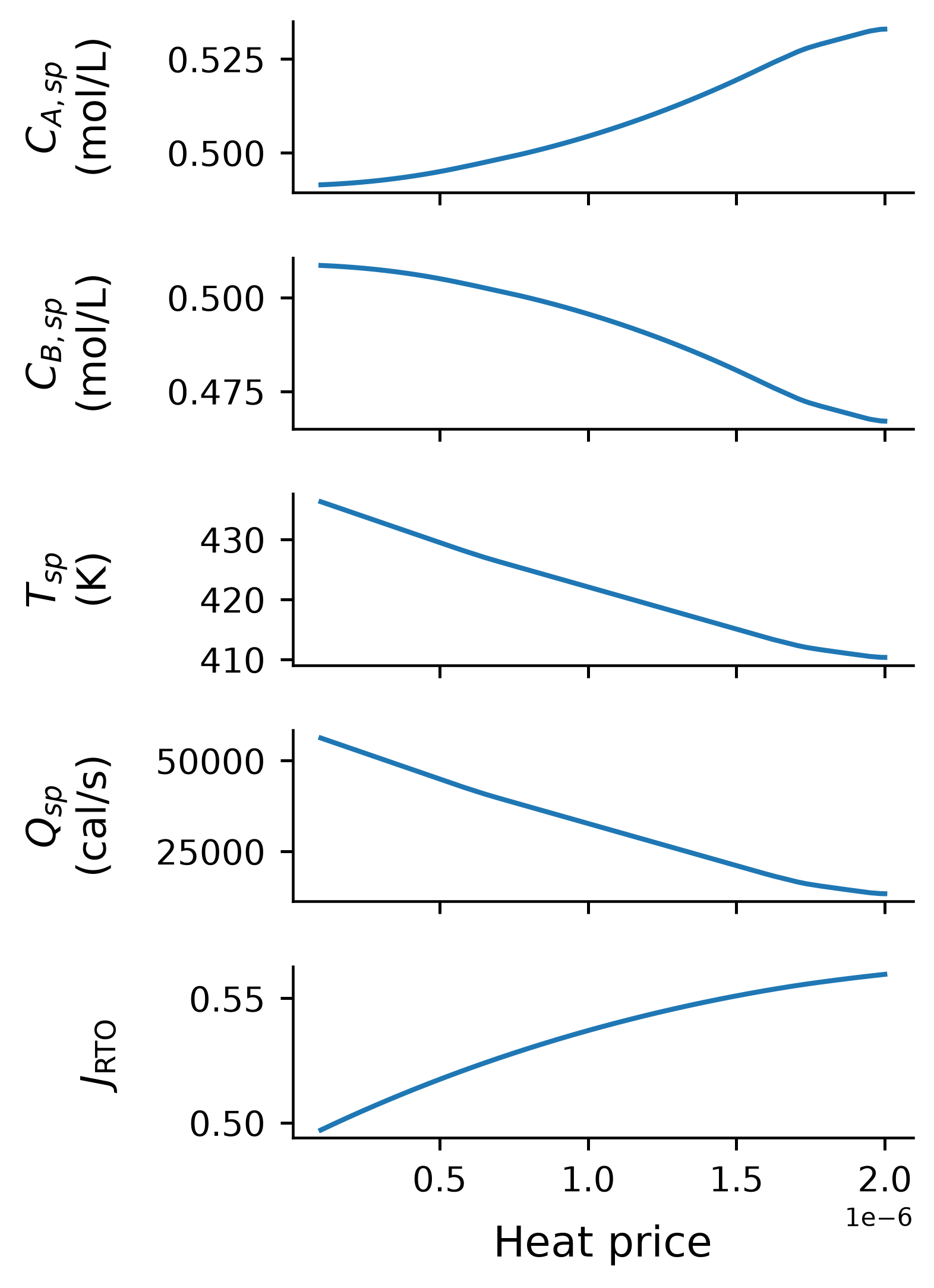}
        \caption{Economic operating map over the complete heat-price range.}
        \label{fig:rto-map}
    \end{subfigure}
    \hfill
    \begin{subfigure}[t]{0.49\textwidth}
        \centering
        \includegraphics[width=0.8\linewidth]{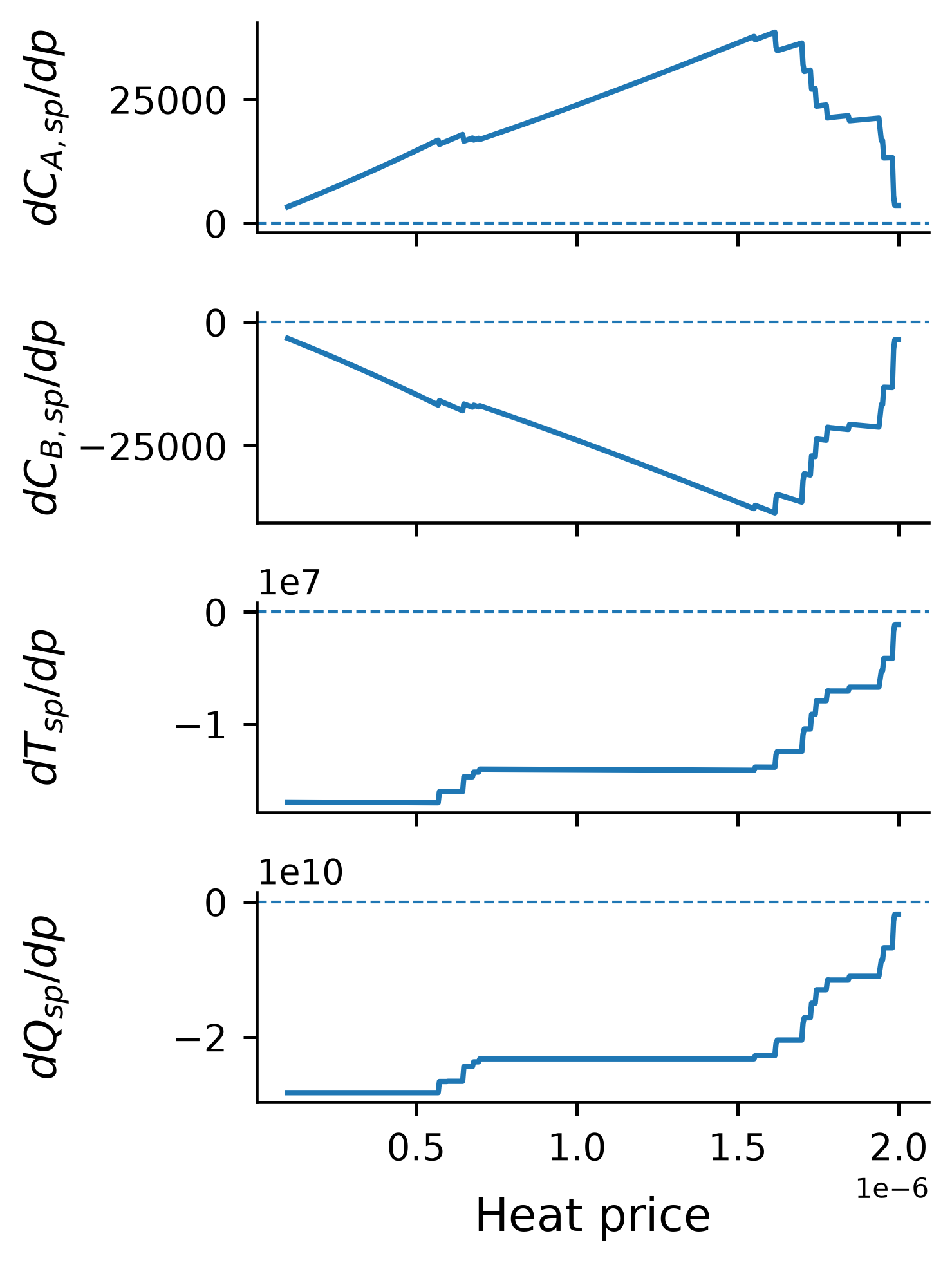}
        \caption{Sensitivity of the optimal operating point with respect to heat price.}
        \label{fig:rto-sensitivity}
    \end{subfigure}
    \caption{Learned RTO policy and its local sensitivity to heat price.}
    \label{fig:rto-map-analysis}
\end{figure}

The learned RTO policy is further evaluated over the complete heat-price range, as shown in Fig.~\ref{fig:rto-map-analysis}. Increasing the heat price reduces the optimal heating rate from approximately $5.62\times10^{4}$ to $1.33\times10^{4}$ cal/s and decreases the optimal reactor temperature from $436.27$ to $410.34$ K. Correspondingly, $C_A^{sp}$ increases from $0.491$ to $0.533$ mol/L and $C_B^{sp}$ decreases from $0.509$ to $0.467$ mol/L, reflecting the trade-off between conversion and energy cost. The optimal objective increases from $0.497$ to $0.560$. The sensitivity profiles show the same trend, with $\mathrm{d}Q^{sp}/\mathrm{d}p<0$, $\mathrm{d}T^{sp}/\mathrm{d}p<0$, $\mathrm{d}C_A^{sp}/\mathrm{d}p>0$, and $\mathrm{d}C_B^{sp}/\mathrm{d}p<0$.

For real-time evaluation, the trained RTO model was tested over $500$ warm single-instance predictions. The mean and median inference latencies were $38.34$ and $38.32$ ms, respectively, with a maximum of $66.49$ ms, which is negligible compared to the $1000$ s RTO update interval. The first model evaluation required $7.67$ s because of one-time JAX compilation and was excluded from the online timing statistics.


\section{PCML-based Model Predictive Control}
\label{sec:cstr-lmpc}
Model predictive control (MPC) determines control actions by repeatedly solving a finite-horizon optimization problem using the current process state while explicitly accounting for system dynamics and operating constraints. At each sampling instance, a sequence of future states and control actions is optimized, the first control action is applied to the process, and the optimization is repeated using the updated state. In general, let $\boldsymbol{x}_k$ denote the current state and other time-varying controller parameters, and $\boldsymbol{y}$ contain the predicted states and control actions over the horizon. The MPC problem can be expressed as $\boldsymbol{y}^{*}(\boldsymbol{x}_k)=\arg\min_{\boldsymbol{y}}J(\boldsymbol{x}_k,\boldsymbol{y})$ subject to $h(\boldsymbol{x}_k,\boldsymbol{y})=0$ and $g(\boldsymbol{x}_k,\boldsymbol{y})\leq0$, where $J$ is the finite-horizon control objective and $h$ and $g$ represent the discrete time governing equations and operating constraints.

\begin{figure}[!htbp]
    \centering
    \includegraphics[width=\textwidth]{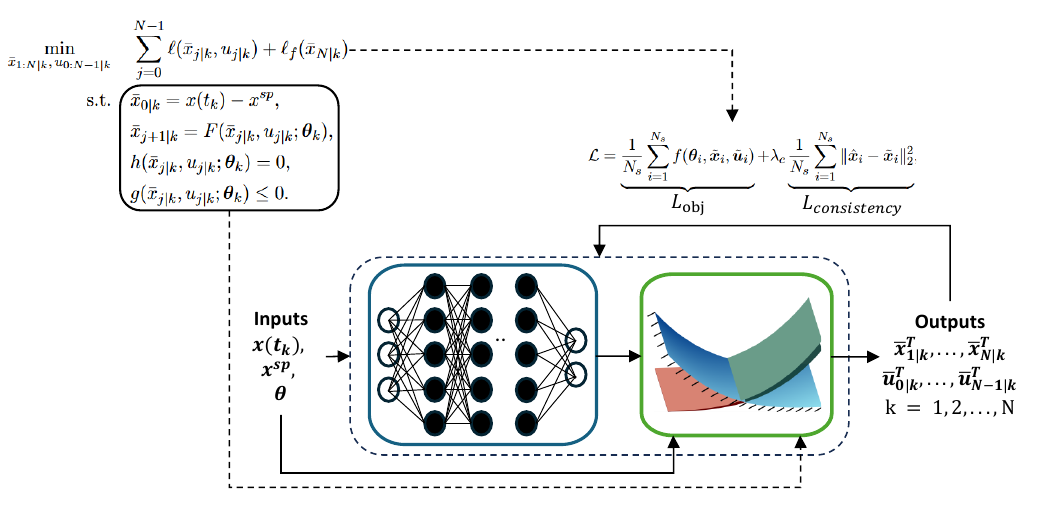}
    \caption{General KKT-HardNet formulation for model predictive control, where the current state, reference, and controller parameters are mapped to the constrained finite-horizon state and control trajectories.}
    \label{fig:mpc_kkthn_diagram}
\end{figure}

Since the optimization must be solved at every sampling instance, the computational cost of nonlinear MPC can become significant for systems with short control intervals or large prediction horizons. KKT-HardNet can instead learn the parametric solution map $\boldsymbol{x}_k\mapsto\boldsymbol{y}^{*}$ offline, while incorporating the MPC constraints through the projection layer. During online operation, the current state and controller parameters are supplied to the trained model to obtain a feasible near-optimal control solution through inference, from which only the first control action is implemented according to the receding-horizon principle. This provides a general approach for accelerating repeated constrained optimization in MPC while retaining the specified process and operating constraints. The general formulation of real-time optimization with KKT-HardNet is illustrated in Figure \ref{fig:mpc_kkthn_diagram}.

\subsection{A CSTR Control Example}
We consider the same CSTR system for the control case study. Given the optimal steady-state condition $(x^{sp},Q^{sp})$, with $x^{sp}=[C_A^{sp},C_B^{sp},T^{sp}]^\top$, the MPC uses the current state $x=[C_A,C_B,T]^\top$ together with these set points to determine the heat input $Q$. 
The MPC is formulated in deviation coordinates, $\bar{x}=x-x^{sp}$ and $u=Q-Q^{sp}$, such that $x=x^{sp}+\bar{x}$ and $Q=Q^{sp}+u$. Defining the net reaction rate as $r=k_Ae^{-E_A/(RT)}C_A-k_Be^{-E_B/(RT)}C_B$, the CSTR dynamics can be written in control-affine form as

\begin{ceqn}
\begin{equation}
\dot{x}=f(x;Q^{sp})+gu,
\label{eq:dynamic_mpc_equality}
\end{equation}
\end{ceqn}

where

\begin{ceqn}
\begin{equation}
f(x;Q^{sp})=
\begin{bmatrix}
\dfrac{C_{A_0}-C_A}{\tau}-r \\
-\dfrac{C_B}{\tau}+r \\
-\dfrac{\Delta H}{\rho C_P}r+\dfrac{T_0-T}{\tau}+\dfrac{Q^{sp}}{\rho C_PV}
\end{bmatrix},
\qquad
g=
\begin{bmatrix}
0\\
0\\
\dfrac{1}{\rho C_PV}
\end{bmatrix}.
\label{eq:lmpc_affine_dynamics}
\end{equation}
\end{ceqn}

To promote stabilizing closed-loop behavior, we consider the Lyapunov function $V(\bar{x})=10^5\bar{C}_A^2+10^5\bar{C}_B^2+\bar{T}^2$, with $L_fV=\nabla V^\top f$ and $L_gV=\nabla V^\top g$. The auxiliary stabilizing control law is

\begin{ceqn}
\begin{equation}
h(\bar{x})=-\frac{\left(L_fV+\sqrt{(L_fV)^2+(L_gV)^4}\right)L_gV}{(L_gV)^2+\epsilon},
\qquad \epsilon=10^{-12}.
\label{eq:auxiliary_control}
\end{equation}
\end{ceqn}

Defining $\dot V(\bar{x},u)=L_fV(\bar{x})+L_gV(\bar{x})u$, the first control action is constrained such that $\dot V(\bar{x}_{0|k},u_{0|k})\leq\dot V(\bar{x}_{0|k},h(\bar{x}_{0|k}))$. The state bounds are $x_{\min}=[0,0,400]^\top$ and $x_{\max}=[1,1,500]^\top$, the heat-input bounds are $Q_{\min}=0$ and $Q_{\max}=10^5$, and consecutive predicted control moves are limited by $\Delta Q_{\max}=5\times10^3$. The resulting parametric MPC problem is as follows:

\begin{ceqn}
\begin{equation}
\begin{aligned}
\min_{\bar{x}_{1:N|k},\,u_{0:N-1|k}} \quad
& \Delta t\sum_{j=1}^{N}\|\bar{x}_{j|k}\|_{Q_x}^{2}
+\Delta t\sum_{j=0}^{N-1}R_u u_{j|k}^{2} \\
\mathrm{s.t.}\quad
& \bar{x}_{0|k}=x(t_k)-x^{sp},\quad
\bar{x}_{j+1|k}=\bar{x}_{j|k}+\Delta t\!\left[f(x^{sp}+\bar{x}_{j|k};Q^{sp})+gu_{j|k}\right] \\
& x_{\min}\leq x^{sp}+\bar{x}_{j|k}\leq x_{\max},\quad
Q_{\min}\leq Q^{sp}+u_{j|k}\leq Q_{\max},\quad
|u_{j|k}-u_{j-1|k}|\leq\Delta Q_{\max} \\
& \dot V(\bar{x}_{0|k},u_{0|k})\leq\dot V(\bar{x}_{0|k},h(\bar{x}_{0|k})).
\end{aligned}
\label{eq:lmpc_formulation}
\end{equation}
\end{ceqn}

The problem is parameterized by $\theta_k=[C_A(t_k),C_B(t_k),T(t_k),C_A^{sp},C_B^{sp},T^{sp},Q^{sp}]^\top$. Only the first optimized control action is implemented, giving $Q(t_k)=Q^{sp}+u_{0|k}^{*}$, after which the process is propagated to the next sampling instance and the optimization is repeated.

We use a sampling time of $5$ s and a prediction/control horizon of $50$ s, corresponding to $N=10$. The resulting NLP contains $7$ parameters, $40$ primal variables, $30$ equality constraints, and $99$ inequality constraints. The $2500$ RTO operating points are used to construct the MPC training conditions with an $80{:}20$ training--validation split. The neural-network backbone contains three hidden layers with $256$ neurons per layer and ReLU activation. The projection uses the primal-dual interior-point method with tolerance $10^{-8}$, maximum $300$ iterations, and barrier regularization parameter $\mu=10^{-4}$. The model is trained for $1200$ epochs using a batch size of $50$, learning rate $3\times10^{-4}$, and primal consistency weight $0.01$. The implementation is shown in Code~\ref{lst:lmpc-pseudocode}.

\begin{lstlisting}[
    style=manuscriptcode,
    language=Python,
    caption={Code for training the Model Predictive Control (MPC) model with KKT-HardNet.},
    label={lst:lmpc-pseudocode}
]
import os
os.environ["JAX_ENABLE_X64"] = "True"
from pathlib import Path
import jax
jax.config.update("jax_enable_x64", True)
import jax.numpy as jnp
from kkthn import KKTHardNet, Expression, Loss, MLP, Projection

SEED = 42, T_FEED = 400.0, CA_FEED = 1.0, TAU = 60.0, KA = 5000.0, KB = 1.0e6, 
EA = 1.0e4, EB = 1.5e4, R = 1.987, DEL_H = -5000.0, RHO = 1.0, CP = 1000.0, V = 100.0
DT = 5.0, N = 10, QC_CA = 1.0, QC_CB = 1.0, QC_T = 5e-5, RC = 1e-11, 
Q_MIN = 0.0, Q_MAX = 1.0e5, DU_MAX = 5.0e3

PARAMETER_NAMES = ["C_A0", "C_B0", "T0", "C_Asp", "C_Bsp", "Tsp", "Qsp"]
STATE_VARIABLE_NAMES = [f"x{v}_{k}" for k in range(1, N+1) for v in ("CA", "CB", "T")]
CONTROL_VARIABLE_NAMES = [f"u{k}" for k in range(N)]
VARIABLE_NAMES=STATE_VARIABLE_NAMES + CONTROL_VARIABLE_NAMES

def objective(theta, y):
    return DT * sum(QC_CA*y[3*k]**2 + QC_CB*y[3*k+1]**2 + QC_T*y[3*k+2]**2 + RC*y[30+k]**2 for k in range(N))

def equalities(theta, y):
    CA0, CB0, T0, CAsp, CBsp, Tsp, Qsp = theta
    residuals = []
    for k in range(N):
        prev = ((CA0 - CAsp, CB0 - CBsp, T0 - Tsp) if k == 0 else y[3 * (k - 1):3 * k])
        xCA_prev, xCB_prev, xT_prev = prev
        xCA_next, xCB_next, xT_next = y[3 * k:3 * k + 3]
        u = y[30 + k]
        CA = CAsp + xCA_prev
        CB = CBsp + xCB_prev
        T = Tsp + xT_prev
        Q = Qsp + u
        r = (KA * jnp.exp(-EA / (R * T)) * CA - KB * jnp.exp(-EB / (R * T)) * CB)
        residuals += [xCA_next - xCA_prev - DT * ((CA_FEED - CA) / TAU - r),
                      xCB_next - xCB_prev - DT * (-CB / TAU + r),
        xT_next - xT_prev - DT*((-DEL_H/(RHO*CP))*r + (T_FEED-T)/TAU + Q/(RHO*CP*V))]
    return jnp.stack(residuals)

def inequalities(theta, y):
    CA0, CB0, T0, CAsp, CBsp, Tsp, Qsp = theta
    xCA0, xCB0, xT0 = CA0 - CAsp, CB0 - CBsp, T0 - Tsp
    rA0 = KA * jnp.exp(-EA / (R * T0)) * CA0
    rB0 = KB * jnp.exp(-EB / (R * T0)) * CB0
    f1 = (CA_FEED - CA0) / TAU - rA0 + rB0
    f2 = -CB0 / TAU + rA0 - rB0
    f3 = (-DEL_H / (RHO * CP)) * (rA0 - rB0) + (T_FEED - T0) / TAU + Qsp / (RHO * CP * V)
    V_CA, V_CB, V_T = 2.0e5 * xCA0, 2.0e5 * xCB0, 2.0 * xT0
    LfV = V_CA * f1 + V_CB * f2 + V_T * f3
    LgV = V_T / (RHO * CP * V)
    h = -(LfV + jnp.sqrt(LfV**2 + LgV**4)) * LgV / (LgV**2 + 1e-12)
    residuals = [LfV + LgV * y[30] - (LfV + LgV * h)]
    for k in range(N):
        xCA, xCB, xT = y[3*k:3*k+3]
        u = y[30+k]
        residuals += [-xCA-CAsp, xCA+CAsp-1, -xCB-CBsp, xCB+CBsp-1, 400-Tsp-xT, 
                      xT+Tsp-500, Q_MIN-Qsp-u, u+Qsp-Q_MAX]
        if k:
            du = u-y[30+k-1]
            residuals += [du-DU_MAX, -du-DU_MAX]
    return jnp.stack(residuals)

def _theta_from_ctx(ctx): 
    return jnp.stack([ctx.parameters[ctx.parameter_index[n]] for n in PARAMETER_NAMES])
def lmpc_primal_transform(raw_y, X):
    temperature_indices = jnp.asarray([2, 5, 8, 11, 14, 17, 20, 23, 26, 29])
    Tsp = X[..., 5:6]
    T = Tsp + raw_y[..., temperature_indices]
    xT = 400.0 + jax.nn.relu(T - 400.0) - Tsp
    return raw_y.at[..., temperature_indices].set(xT)
    
# Configurations
TRAIN = {"epochs": 1200, "batch_size": 50, "learning_rate": 3e-4, "train_frac": 0.8, 
         "seed": SEED, "dtype": "float64", "print_every": 1, "standardize_inputs": True,
         "pretrain": False, "drop_last_batch": True, "backend": "cpu", 
         "output_dir": str(Path.cwd()/"runs")}
DATA_PATH = Path.cwd()/"lmpc"/"parameters_new.csv"

# Backbone
nn = MLP(hidden_sizes=(256, 256, 256), activation="relu", 
         predict_duals=True, inequality_dual_floor=1e-8)

# Projection Layer
projection = Projection(name="CSTR_LMPC", method="primal-dual-interior-point", 
                        config={"tolerance": 1e-8,"max_iterations": 300,"mu": 1e-4})
projection.add_parameter(PARAMETER_NAMES)
projection.add_variable(VARIABLE_NAMES)

equality_dependencies, inequality_dependencies = [], [("u0",)]
for k in range(N):
    previous = [] if k == 0 else [f"xCA_{k}", f"xCB_{k}", f"xT_{k}"]
    for v in ("xCA", "xCB", "xT"): 
        equality_dependencies.append(
        tuple(previous + [f"{v}_{k+1}"] + ([f"u{k}"] if v == "xT" else [])))
    for v in ("xCA", "xCB", "xT"):
        inequality_dependencies += [(f"{v}_{k+1}",), (f"{v}_{k+1}",)]
    inequality_dependencies += [(f"u{k}",), (f"u{k}",)]
    if k > 0:
        inequality_dependencies += [(f"u{k-1}", f"u{k}"),(f"u{k-1}", f"u{k}")]

for i, deps in enumerate(equality_dependencies):
    expr = Expression(
        lambda ctx, i=i:equalities(_theta_from_ctx(ctx), ctx.variables)[i],f"h[{i}]",
        variable_dependencies=deps,parameter_dependencies=PARAMETER_NAMES,
        affine_in_variables=False,)
    projection.constraints.equality(expr)

for i, deps in enumerate(inequality_dependencies):
    expr = Expression(
        lambda ctx, i=i:inequalities(_theta_from_ctx(ctx), ctx.variables)[i],f"g[{i}]",
        variable_dependencies=deps,parameter_dependencies=PARAMETER_NAMES,
        affine_in_variables=True,)
    projection.constraints.inequality(expr)

projection.objective = Expression(
    lambda ctx: objective(_theta_from_ctx(ctx),ctx.variables,),"objective",
    variable_dependencies=VARIABLE_NAMES, parameter_dependencies=PARAMETER_NAMES,
    affine_in_variables=False,)    
# Loss function
loss = Loss.objective(primal_consistency_weight=0.01, equality_dual_consistency_weight=0.0,
                        inequality_dual_consistency_weight=0.0)

# Model definition
model = KKTHardNet("optimization", nn, loss, projection, train=TRAIN)
model.primal_transform_fn = lmpc_primal_transform
model.dataset(parameters=DATA_PATH)

# Training
pdip_result = model.optimize()

# Inference
model = KKTHardNet().load(Path("/path/to/metadata.json"))
x_new = jnp.array([[0.50, 0.45, 425.0, 0.48, 0.47, 426.0, 40000]])
y_pred = model.predict(x_new)
\end{lstlisting}

\begin{table}[!htbp]
\caption{Performance comparison for the CSTR MPC problem in terms of objective value, constraint satisfaction, training time, and inference time for a batch of $100$ instances.}
\label{tab:lmpc-performance}
\centering
\resizebox{\textwidth}{!}{%
\begin{tabular}{lrrrrrrr}
\toprule
 & Objective & Mean eq. & Max eq. & Mean ineq. & Max ineq. & Training Time (s) & Inference Time (ms/batch) \\
\midrule
IPOPT & $1.32\times10^{-1}$ & $4.75\times10^{-10}$ & $5.63\times10^{-8}$ & $4.03\times10^{-14}$ & $9.98\times10^{-9}$ & $--$ & $6.30^*$ \\
PINN & $1.19\times10^{-1}$ & $5.31\times10^{-3}$ & $8.16\times10^{-2}$ & $6.18\times10^{-6}$ & $2.72\times10^{-2}$ & $137.68$ & $0.72$ \\
DC3 & $2.70\times10^{-1}$ & $0.00$ & $0.00$ & $1.39\times10^{3}$ & $3.45\times10^{4}$ & $1821.28$ & $31.60$ \\
KKT-HardNet & $1.60\times10^{-1}$ & $2.28\times10^{-5}$ & $1.48\times10^{-3}$ & $0.00$ & $0.00$ & $10364.44$ & $166.07$ \\
\bottomrule
\end{tabular}
}
\footnotesize
*Solver time is reported as average solve time considering full parallelization across 100 instances
\end{table}

As shown in Table~\ref{tab:lmpc-performance}, KKT-HardNet obtains an average objective value of $1.60\times10^{-1}$ compared to $1.32\times10^{-1}$ obtained by the reference optimizer, while maintaining zero inequality violation and small equality residuals. The PINN obtains a lower objective value but remains infeasible, whereas DC3 satisfies the equality constraints but exhibits large inequality violations. The reported inference times correspond to batches of $100$ instances; however, closed-loop control requires sequential single-instance evaluations, which are analyzed separately below.

\begin{figure}[!htbp]
    \centering
    \includegraphics[width=0.75\textwidth]{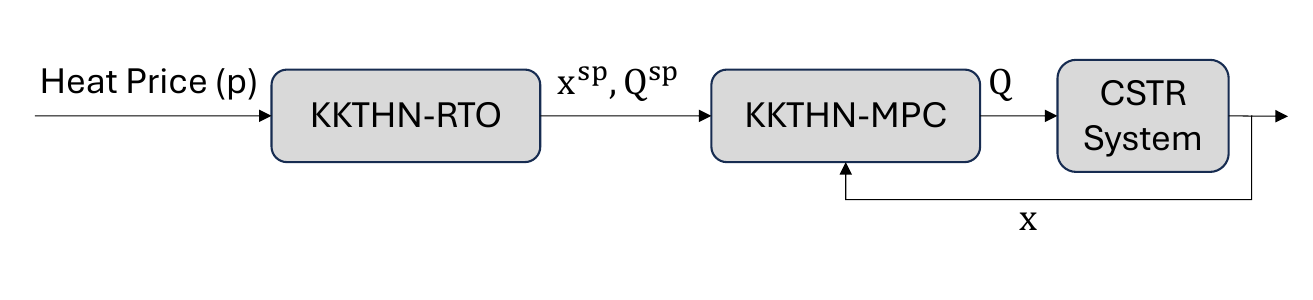}
    \caption{Closed-loop RTO--MPC architecture for the nonlinear CSTR. The RTO layer provides the economically optimal operating target, while the MPC regulates the process toward the updated set point subject to operating and Lyapunov-based constraints.}
    \label{fig:mpc-schematic}
\end{figure}

\begin{figure}[!htbp]
    \centering
    \includegraphics[width=0.8\textwidth]{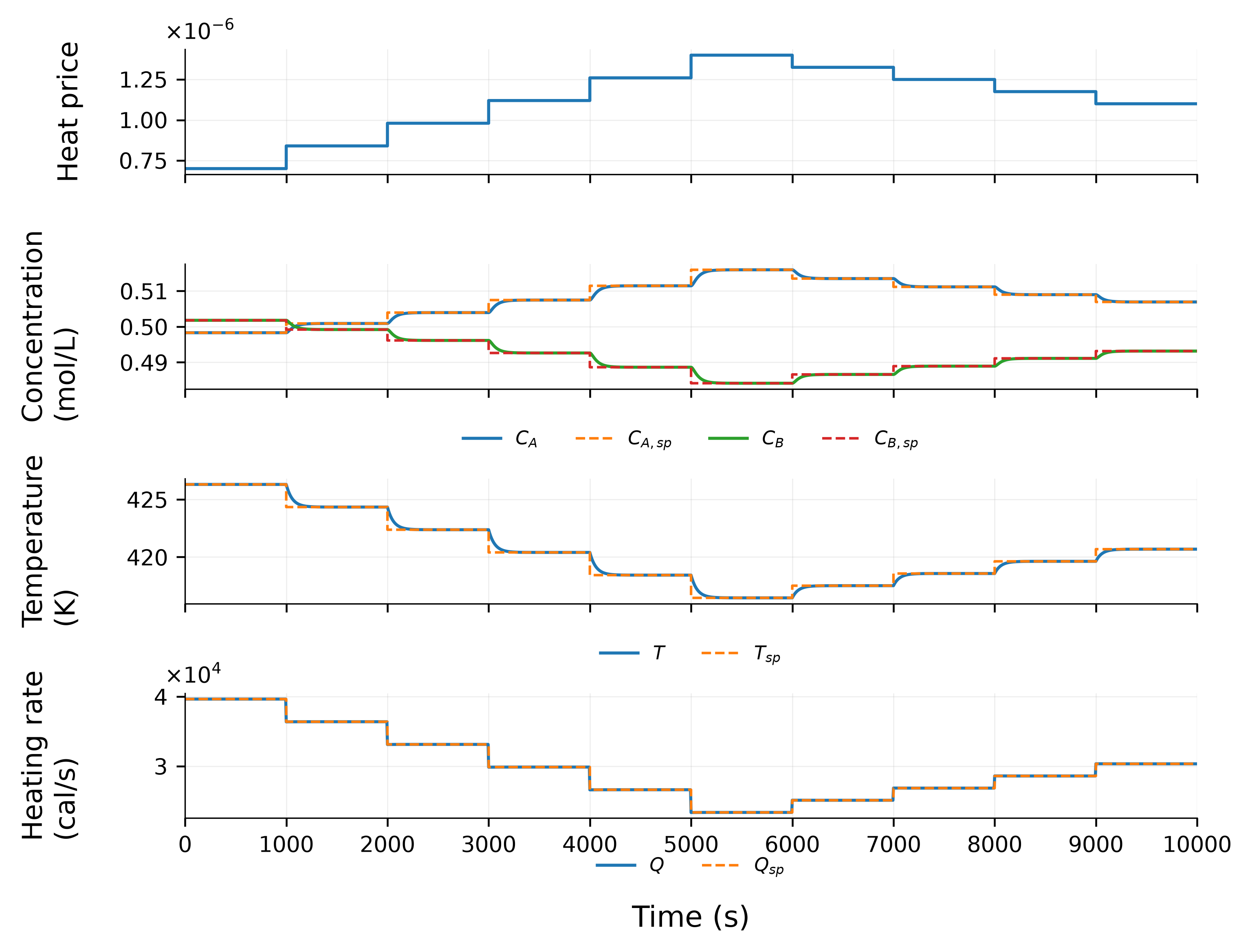}
    \caption{Closed-loop RTO--MPC response under the time-varying heat-price profile. The RTO layer updates the economically optimal set points $(C_A^{sp},C_B^{sp},T^{sp},Q^{sp})$, while the KKT-HardNet MPC regulates the nonlinear CSTR toward the updated operating conditions.}
    \label{fig:rto-lmpc-closed-loop}
\end{figure}

We integrate the previous RTO model, which finds the optimal set points, with the MPC model to complete the closed loop control, as illustrated in Figure ~\ref{fig:mpc-schematic}. The economically optimal operating targets obtained from the RTO layer are used as set points for the MPC. The closed-loop RTO--MPC response is shown in Fig.~\ref{fig:rto-lmpc-closed-loop}. The RTO layer is updated every $1000$ s, while the MPC is evaluated every $5$ s. A perfect actuator is assumed \cite{zhang2019real}, and the nonlinear CSTR is propagated using fourth-order Runge--Kutta integration. The updated state is supplied to the controller at the next sampling instance, forming the receding-horizon closed loop.

\begin{table}[!htbp]
\caption{Closed-loop tracking performance of the KKT-HardNet MPC controller. Settling time is defined as the time required to enter and remain within $10^{-4}$ mol/L of the concentration set points and $0.05$ K of the temperature set point until the subsequent RTO update.}
\label{tab:lmpc-tracking}
\centering
\begin{tabular}{lrrrr}
\toprule
Variable & RMSE & Mean settling time (s) & Max. overshoot (\%) & Mean steady-state abs. error \\
\midrule
$C_A$ & $6.35\times10^{-4}$ & $222.8$ & $0.026$ & $5.27\times10^{-7}$ \\
$C_B$ & $6.34\times10^{-4}$ & $223.3$ & $0.014$ & $2.85\times10^{-7}$ \\
$T$ & $0.278$ & $204.4$ & $0.004$ & $3.47\times10^{-5}$ \\
\bottomrule
\end{tabular}
\end{table}

Figure~\ref{fig:rto-lmpc-closed-loop} and Table~\ref{tab:lmpc-tracking} illustrate that the controller closely tracks the economically optimal set points. The RMSE values for $C_A$, $C_B$, and $T$ are $6.35\times10^{-4}$ mol/L, $6.34\times10^{-4}$ mol/L, and $0.278$ K, respectively. Following each RTO update, the process settles within approximately $185$--$255$ s, well within the $1000$ s RTO interval, with maximum overshoot below $0.03\%$ for all controlled variables.

Constraint satisfaction was verified over all $2000$ controller evaluations. The mean and maximum absolute equality residuals were $1.90\times10^{-9}$ and $5.15\times10^{-7}$, respectively, while the mean and maximum inequality violations were $7.72\times10^{-13}$ and $2.95\times10^{-9}$. No violations were observed for the predicted state bounds, heat-input bounds, horizon control-move constraints, or simulated plant operating bounds. The remaining small inequality residuals arise from the Lyapunov condition and remain below the $10^{-8}$ projection tolerance.

\begin{figure}[!htbp]
    \centering
    \includegraphics[width=0.8\textwidth]{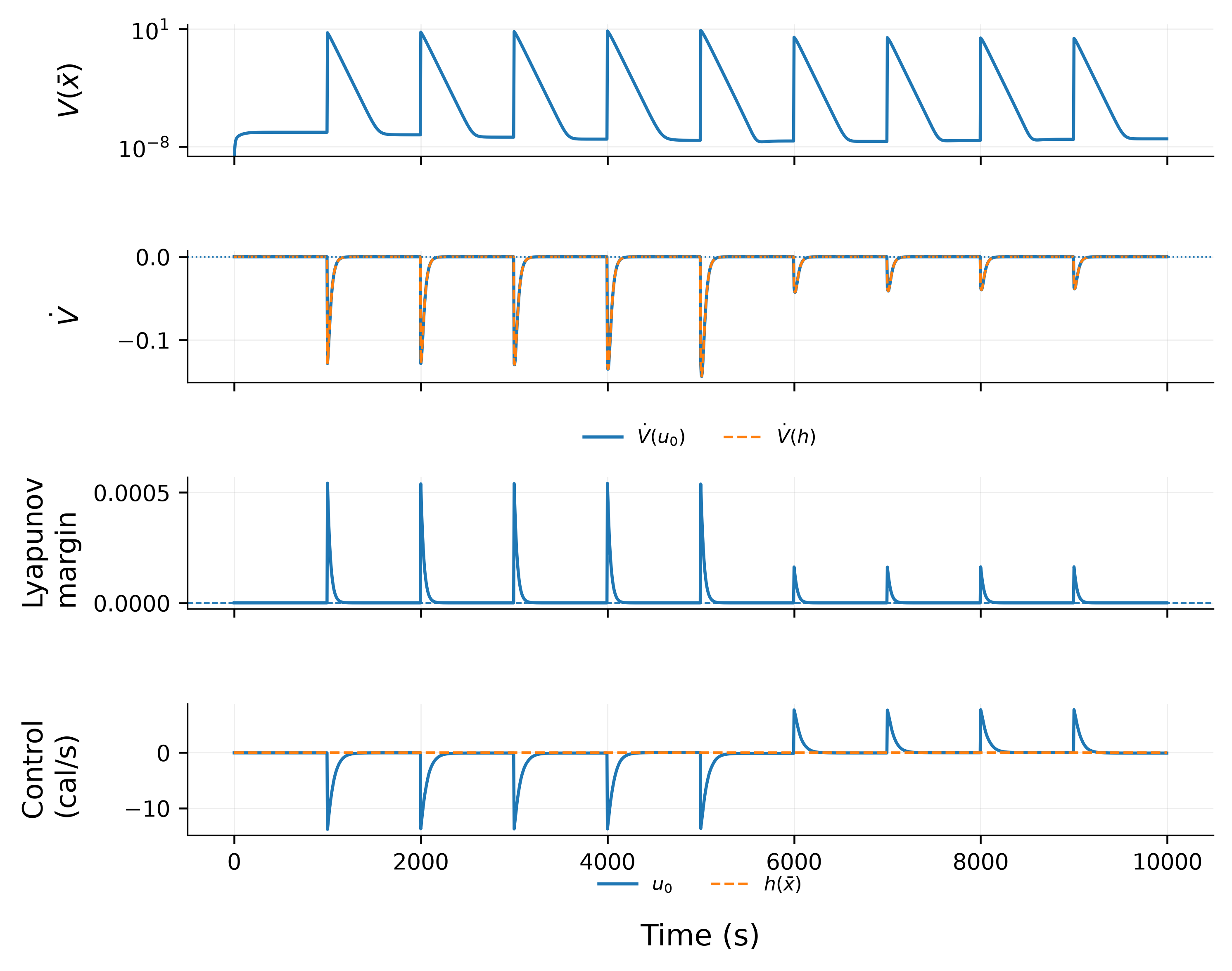}
    \caption{Closed-loop Lyapunov analysis of the KKT-HardNet MPC controller. The Lyapunov function increases after each RTO set-point update because the reference changes and subsequently decays toward zero as the process tracks the new target. The Lyapunov margin remains non-negative up to numerical tolerance.}
    \label{fig:lmpc-lyapunov}
\end{figure}

The Lyapunov based analysis in Fig.~\ref{fig:lmpc-lyapunov} shows that $V(\bar{x})$ increases immediately after each reference update and subsequently decays toward zero. Across all $2000$ evaluations, the condition $\dot V(\bar{x},u_0)\leq\dot V(\bar{x},h(\bar{x}))$ exhibits no violations above the $10^{-8}$ tolerance. The minimum Lyapunov margin is $-2.95\times10^{-9}$ and the mean margin is $1.19\times10^{-5}$, indicating consistent satisfaction of the imposed Lyapunov condition. This behavior shows that the controller drives the system toward the updated equilibrium after each set-point change, providing numerical evidence of closed-loop convergence and stability. The consistent satisfaction of the Lyapunov condition further supports the stabilizing behavior of the controller.

The real-time computational performance of the MPC was evaluated over $2000$ warm single-instance predictions. The mean and median inference latencies were $20.20$ and $19.94$ ms, while the $95$th and $99$th percentile latencies were $29.92$ and $34.86$ ms. The maximum latency was $79.41$ ms, corresponding to only $1.59\%$ of the $5$ s sampling interval, and no deadline misses were observed. The first evaluation required approximately $35.95$ s because of one-time JAX compilation and was excluded from the online timing statistics. At the nine instances where both RTO and MPC were evaluated sequentially, the combined median and maximum latencies were $50.95$ and $59.48$ ms, respectively.

\begin{figure}[!htbp]
    \centering
    \includegraphics[width=0.8\textwidth]{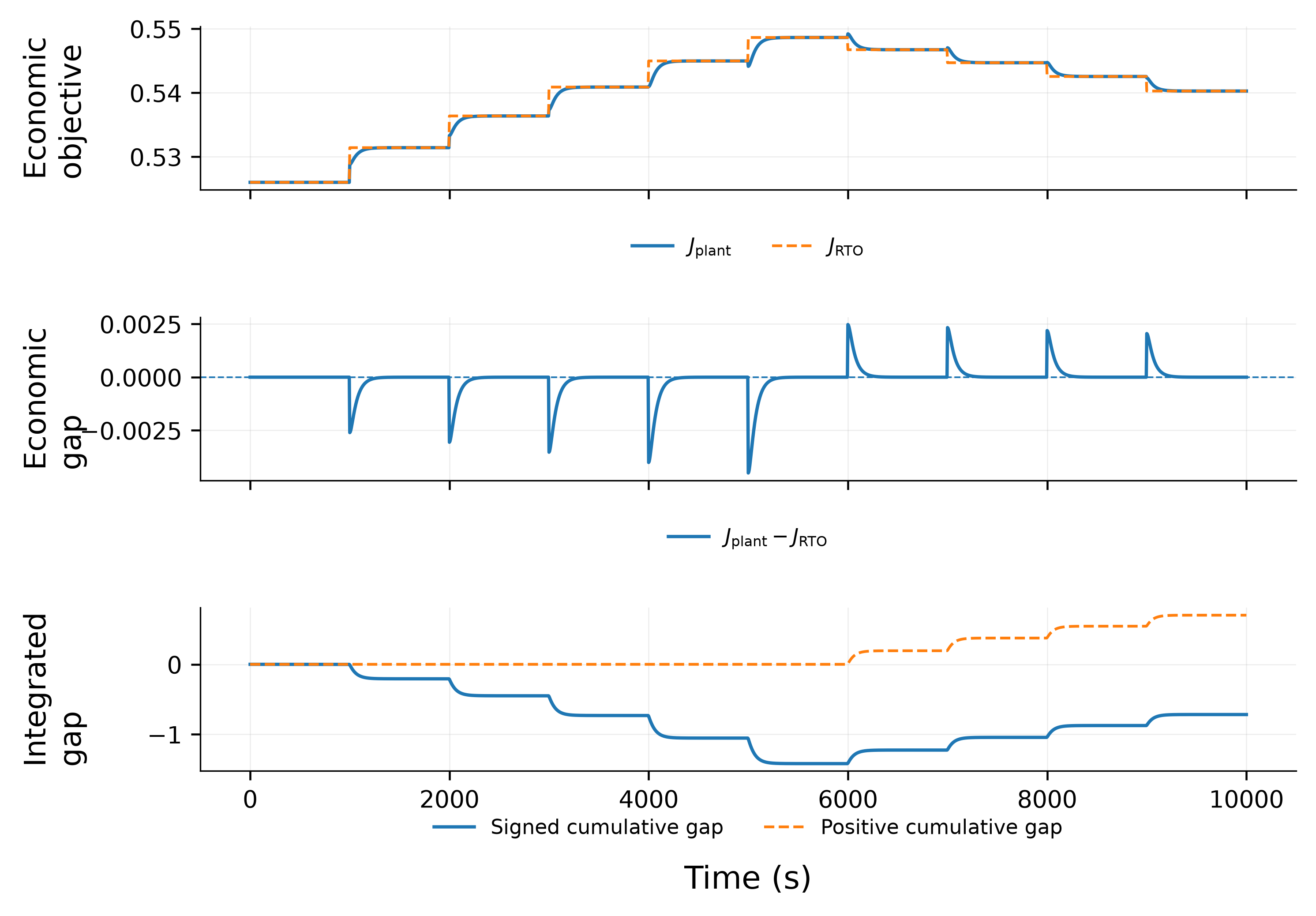}
    \caption{Closed-loop economic performance of the coupled RTO--MPC framework. The top panel compares the instantaneous plant objective $J_{\mathrm{plant}}$ with the steady-state RTO target $J_{\mathrm{RTO}}$, while the remaining panels show the instantaneous and cumulative economic gaps.}
    \label{fig:rto-lmpc-economic-performance}
\end{figure}

The closed-loop economic performance is evaluated using $J_{\mathrm{plant}}=C_A/C_{A_0}+pQ$ and $J_{\mathrm{RTO}}=C_A^{sp}/C_{A_0}+pQ^{sp}$. Over the complete $10{,}000$ s simulation, the mean plant and RTO objectives were $0.54018$ and $0.54025$, respectively, with a mean absolute economic gap of $2.12\times10^{-4}$. The maximum absolute gap was $4.50\times10^{-3}$ and occurred immediately after an RTO update, after which the MPC rapidly returned the process objective toward the RTO target. Additional analyses of control effort, robustness, alternative heat-price profiles, and local controller sensitivity are provided in Appendix~\ref{app:additional-control-analysis}.

\section{PCML for Grey-Box Optimization} \label{sec:syngas}
Data-driven optimization is useful when the relationship between operating conditions and process performance is available primarily through experiments, sensor measurements, or high-fidelity simulations rather than an explicit differentiable process model. A surrogate model can first be trained to approximate this input--output relationship and subsequently used as an efficient model for optimization. In general, let $\boldsymbol{x}\in\mathcal{X}$ denote the controllable operating conditions and $\boldsymbol{y}=\mathcal{M}(\boldsymbol{x})$ the outputs predicted by a trained surrogate $\mathcal{M}$. The downstream optimization problem can then be expressed as $\boldsymbol{x}^{*}=\arg\min_{\boldsymbol{x}\in\mathcal{X}}F(\boldsymbol{x},\mathcal{M}(\boldsymbol{x}))$, where $F$ represents the desired process-performance or economic objective.

The reliability of this approach depends on the accuracy and physical consistency of the learned surrogate, since the optimizer may otherwise identify operating conditions that exploit prediction errors or produce physically inconsistent outputs. KKT-HardNet provides a differentiable surrogate whose predictions satisfy the physical constraints incorporated during training. Consequently, gradients of the downstream objective can be propagated through the trained KKT-HardNet model to optimize the input conditions directly, while the surrogate predictions remain constrained by the available domain knowledge. This provides a general framework for using observation-based models not only for prediction, but also for identifying improved operating conditions.

\subsection{Simulation-based Optimization of a Syngas Production Process}
\begin{figure}[ht]
    \centering
    \includegraphics[width=0.95\textwidth]{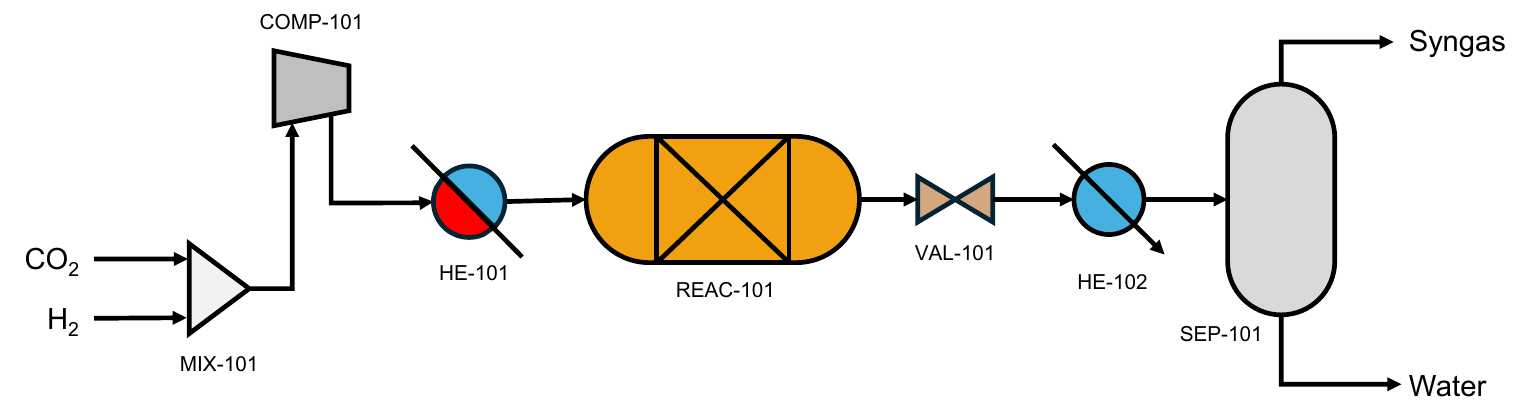}
    \caption{A syngas production pathway}
    \label{fig:syngas_flowsheet}
\end{figure}

We consider the syngas production pathway illustrated in Fig.~\ref{fig:syngas_flowsheet}. The process uses $\mathrm{CO_2}$ and $\mathrm{H_2}$ as feed streams, both entering the mixer at a temperature of $25^{\circ}\mathrm{C}$ and a pressure of $1~\mathrm{atm}$. The $\mathrm{CO_2}$ feed flow rate is fixed at $1~\mathrm{kmol,h^{-1}}$ as the basis, while the $\mathrm{H_2}$ feed flow rate is allowed to vary. We define the ratio between the $\mathrm{CO_2}$ and $\mathrm{H_2}$ feed flow rates as $a=F_{\mathrm{CO_2}}/F_{\mathrm{H_2}}$, where $F_{\mathrm{CO_2}}$ and $F_{\mathrm{H_2}}$ denote the molar flow rates of $\mathrm{CO_2}$ and $\mathrm{H_2}$, respectively.

The two feed streams are first combined in a mixer, after which the mixed gas stream is compressed from the feed pressure to the desired operating pressure, $P_{\mathrm{op}}$, using an isentropic compressor. The compressor is driven by an electrical utility. Following compression, a simple heat exchanger is used to bring the mixed gas stream to the desired reactor operating temperature, $T_{\mathrm{op}}$. Depending on the temperature of the compressed stream relative to the specified reactor operating temperature, the heat exchanger can provide either heating or cooling. The heating requirement can be supplied using a heating utility, such as steam or electricity, whereas the cooling requirement can be supplied using a cooling utility, such as cooling water or refrigerant. In this manner, the feed stream entering the reactor is maintained at the specified operating temperature $T_{\mathrm{op}}$ and operating pressure $P_{\mathrm{op}}$. The conditioned mixed-gas stream is subsequently introduced into a Gibbs equilibrium reactor. Instead of employing an explicit kinetic reaction model, the equilibrium reactor determines the product composition by minimizing the total Gibbs free energy of the system subject to elemental conservation constraints. The chemistry occurring in the reactor can primarily be described by the $\mathrm{CO_2}$ methanation and reverse water--gas shift reactions,

\begin{ceqn}
\begin{equation}
\mathrm{CO_2 + 4H_2 \rightleftharpoons CH_4 + 2H_2O},
\end{equation}
\end{ceqn}

\begin{ceqn}
\begin{equation}
\mathrm{CO_2 + H_2 \rightleftharpoons CO + H_2O}.
\end{equation}
\end{ceqn}

The Gibbs equilibrium reactor is configured with a maximum of $50$ equilibrium iterations. The reactor therefore predicts the equilibrium distribution among the species $\mathrm{CO_2}$, $\mathrm{H_2}$, $\mathrm{CO}$, $\mathrm{CH_4}$, and $\mathrm{H_2O}$ at the specified operating conditions $T_{\mathrm{op}}$ and $P_{\mathrm{op}}$. Following the reactor, the product stream is passed through a valve, where it is depressurized from the reactor operating pressure to approximately $1~\mathrm{atm}$. The depressurized stream is subsequently passed through a cooler, where its temperature is reduced to $50^{\circ}\mathrm{C}$. The cooled reactor effluent is then introduced into a vapor--liquid separator operating at approximately $T_{\mathrm{sep}}=25^{\circ}\mathrm{C}$ and $1~\mathrm{atm}$. The separator performs vapor--liquid equilibrium separation and produces two outlet streams. The top stream corresponds to the gas-phase product, referred to as the syngas stream, and contains five gas-phase components: $\mathrm{CO_2}$, $\mathrm{H_2}$, $\mathrm{CO}$, $\mathrm{CH_4}$, and $\mathrm{H_2O}$. The bottom stream corresponds to the liquid-phase product and primarily contains condensed water.

The Aspen Plus simulation employs the Peng--Robinson equation of state as the thermodynamic property method for evaluating the required phase and equilibrium properties throughout the flowsheet. To generate the dataset used in this work, we sample $4000$ operating points by varying the feed-flow ratio $a$, reactor operating temperature $T_{\mathrm{op}}$, and reactor operating pressure $P_{\mathrm{op}}$. The operating points are generated using Latin hypercube sampling with a random seed of $42$. The corresponding sampling ranges are $a\in[1,10]$, $T_{\mathrm{op}}\in[100,1000]^{\circ}\mathrm{C}$, and $P_{\mathrm{op}}\in[1,20]~\mathrm{atm}$. For each sampled combination of $a$, $T_{\mathrm{op}}$, and $P_{\mathrm{op}}$, the complete Aspen Plus flowsheet is simulated to obtain the corresponding equilibrium product composition and process utility requirements.

\textbf{Prediction Model: }The syngas formulation has $3$ parameters, $6$ primal variables, $3$ equality constraints, and no inequality constraints. The input parameters are the feed ratio $a$, operating pressure $P_{\mathrm{op}}$, and operating temperature $T_{\mathrm{op}}$, while the predicted variables are the outlet molar flows of $\mathrm{CO_2}$, $\mathrm{H_2}$, $\mathrm{CO}$, $\mathrm{CH_4}$, and $\mathrm{H_2O}$, together with the total utility cost. The three equality constraints represent the elemental carbon, oxygen, and hydrogen balances and are given as 
\begin{ceqn}
    \begin{equation}
        \begin{aligned}
            &F_{\mathrm{CO_2}}+F_{\mathrm{CO}}+F_{\mathrm{CH_4}}=1 \\
            &2F_{\mathrm{CO_2}}+F_{\mathrm{CO}}+F_{\mathrm{H_2O}}=2 \\
            &4F_{\mathrm{CH_4}}+2F_{\mathrm{H_2}}+2F_{\mathrm{H_2O}}=2a.
        \end{aligned}
    \end{equation}
\end{ceqn}


We use the $4000$ samples generated from the Aspen Plus simulations and split the dataset in an $80:20$ ratio for training and validation respectively. The network structure consists of $2$ hidden layers with $64$ hidden nodes in each layer and ReLU activation. Since the formulation contains only linear equality constraints, we consider the ``linear-equality" projection formulation \cite{chen2024physics}, where the network predicts the $6$ primal variables together with $3$ equality dual variables. We first perform MLP pretraining for $100$ epochs and subsequently activate the hard-constraint projection. The model is trained for a total of $2000$ epochs with a batch size of $50$, learning rate of $10^{-3}$, random seed $42$. The training loss consists of the mean-squared data loss together with consistency losses for the primal and equality-dual variables, for which both consistency weights are set to $1$. Since no inequality constraints are present in this formulation, no inequality-dual variables are required. The linear-equality projection enforces the three elemental balance constraints during the projection stage, resulting in equality-constraint residuals close to machine precision. The code to train the model is given in codeblock \ref{lst:syngas-pseudocode}.

\begin{lstlisting}[
    style=manuscriptcode,
    language=Python,
    caption={Code for training the syngas model with KKT-HardNet.},
    label={lst:syngas-pseudocode}
]
import os
os.environ["JAX_ENABLE_X64"] = "True"

from pathlib import Path
import numpy as np

import jax
jax.config.update("jax_enable_x64", True)
import jax.numpy as jnp

from kkthn import KKTHardNet, Loss, MLP, Projection

SEED = 42
TRAIN = {
    "epochs": 2000,"epoch_mlp": 100,"batch_size": 50,"learning_rate": 1e-3,"train_frac": 0.8,"seed": SEED,"dtype": "float64","print_every": 100,"minimum_projection_convergence": 0.95,"readiness_extra_epochs": 500,"readiness_check_every": 10,"backend": "cpu","output_dir": str(Path.cwd() / "runs"),}

DATA_PATH = Path.cwd() / "syngas" / "syngas_dataset_lhc.csv"

# Backbone
nn = MLP(hidden_sizes=(64, 64), activation="relu", predict_duals=True, inequality_dual_floor=1e-12)

# Projection
projection = Projection(name="syngas_production", method="linear-equality")

x = projection.add_parameter(["a", "P", "T"])
y = projection.add_variable(["CO2", "H2", "CO", "CH4", "H2O", "Util"])
projection.constraints.add(
    y.CO2 + y.CO + y.CH4 == 1,
    2.0 * y.CO2 + y.CO + y.H2O == 2,
    4.0 * y.CH4 + 2.0 * y.H2 + 2.0 * y.H2O == 2.0 * x.a
)

# Loss
loss = Loss.mse(primal_consistency_weight=1.0, equality_dual_consistency_weight=1.0, inequality_dual_consistency_weight=1.0)

# Model Definition
model = KKTHardNet("surrogate", nn, loss, projection, train=TRAIN)
model.dataset(parameters=DATA_PATH, variables=DATA_PATH)

# Training and Summary
linear_equality_result = model.model()
model.summary()
model.plot_history()
\end{lstlisting}

\begin{table}[!htbp]
\caption{Performance comparison for the syngas production surrogate-modeling problem in terms of prediction accuracy, elemental-balance constraint satisfaction, training time, and inference time for a batch of 100 instances.}
\label{tab:syngas-model-comparison}
\centering
\resizebox{\textwidth}{!}{%
\begin{tabular}{lrrrrrrr}
\toprule
 & MSE & Mean eq. & Max eq. & Mean ineq. & Max ineq. & Training Time (s) & Inference Time (ms/batch) \\
\midrule
MLP & $6.7 \times 10^{-5}$ & $5.40 \times 10^{-3}$ & $0.11$ & --  & -- & 92.02  & 0.2  \\
PINN & $1.31 \times 10^{-4}$ & $6.47 \times 10^{-3}$ & $0.05$ & --  & -- & 90.87  & 0.2  \\
ENFORCE & $3.45 \times 10^{-5}$ & $1.31 \times 10^{-7}$ & $3.81 \times 10^{-6}$ & --  & -- & 226.70  & 0.66 \\ 
KKT-HardNet & $4.48 \times 10^{-5}$ & $2.25 \times 10^{-16}$ & $3.55 \times 10^{-15}$ & --  & -- & 60.80  & 1.73 \\ \bottomrule
\end{tabular}
}
\end{table}

Table~\ref{tab:syngas-model-comparison} compares the predictive accuracy, elemental-balance satisfaction, training cost, and inference performance of the surrogate models for the syngas production case study. KKT-HardNet achieves an MSE of $4.48\times10^{-5}$ while satisfying the three elemental-balance equality constraints to essentially machine precision, with mean and maximum residuals of $2.25\times10^{-16}$ and $3.55\times10^{-15}$, respectively. The unconstrained MLP and PINN attain similar prediction errors but exhibit substantially larger equality residuals, whereas ENFORCE also enforces the balances accurately but requires a longer training time. KKT-HardNet therefore combines high predictive accuracy with exact elemental-balance consistency and the lowest training time among the hard-constrained approaches considered, although its inference time is higher than those of the unconstrained and soft-constrained neural-network baselines.


\begin{figure}[!htbp]
\centering
\begin{adjustbox}{max totalsize={0.95\linewidth}{0.70\textheight},center}
\begin{tikzpicture}[
    flow/.style={-{Stealth[length=2.3mm]},thin},
    block/.style={
        draw,
        rounded corners=5pt,
        align=center,
        text width=5.0cm,
        minimum height=9mm,
        inner sep=2pt,
        fill=white
    },
    decision/.style={
        draw,
        diamond,
        aspect=2.8,
        align=center,
        text width=2.6cm,
        inner sep=1.5pt,
        fill=white
    },
    terminal/.style={
        block,
        fill=gray!20
    },
    label/.style={
        font=\scriptsize,
        fill=white,
        inner sep=1pt,
        align=center
    }
]
\node[terminal] (input) at (0,0)
{Trained surrogate $\mathcal{M}$ from KKT-HardNet and optimization parameters};

\node[block] (alpha) at (0,-1.4)
{Select $\alpha$ and set $F^*\leftarrow+\infty$};

\node[block] (init) at (0,-2.9)
{Sample $(a,P_{\mathrm{op}},T_{\mathrm{op}})$\\
and initialize Adam};

\node[block] (adam) at (0,-5.0)
{Evaluate $\mathcal{M}$ and objective\\
$F=\alpha U-(1-\alpha)X_{\mathrm{CO_2}}$\\
Compute gradient, Adam step, and projection};

\node[decision] (conv) at (0,-7.6)
{Converged or\\$k=K$?};

\node[block] (best) at (0,-10.0)
{If $F<F^*$, update and store best solution};

\node[decision] (starts) at (0,-12.4)
{More random\\starts?};

\node[decision] (alphas) at (0,-14.9)
{More $\alpha$\\values?};

\node[terminal] (output) at (0,-17.2)
{Return optimal operating conditions\\
for each $\alpha$};

\draw[flow] (input.south) -- (alpha.north);
\draw[flow] (alpha.south) -- (init.north);
\draw[flow] (init.south) -- (adam.north);
\draw[flow] (adam.south) -- (conv.north);

\draw[flow]
(conv.east)
-- node[label,above]{No} (6.2,-7.6)
-- node[label,right]{Next iteration} (6.2,-5.0)
-- (adam.east);

\draw[flow]
(conv.south)
-- node[label,right]{Yes} (best.north);

\draw[flow]
(best.south) -- (starts.north);

\draw[flow]
(starts.east)
-- node[label,above]{Yes} (7.4,-12.4)
-- node[label,right]{Next start} (7.4,-2.9)
-- (init.east);

\draw[flow]
(starts.south)
-- node[label,right]{No} (alphas.north);

\draw[flow]
(alphas.west)
-- node[label,above]{Yes} (-6.2,-14.9)
-- node[label,left]{Next $\alpha$} (-6.2,-1.4)
-- (alpha.west);

\draw[flow]
(alphas.south)
-- node[label,right]{No} (output.north);

\end{tikzpicture}
\end{adjustbox}
\caption{Multi-start projected Adam for syngas operating-condition optimization. More details about the algorithm is given in Appendix \ref{app:projected-adam}}
\label{fig:syngas_projected_adam_flowchart}
\end{figure}
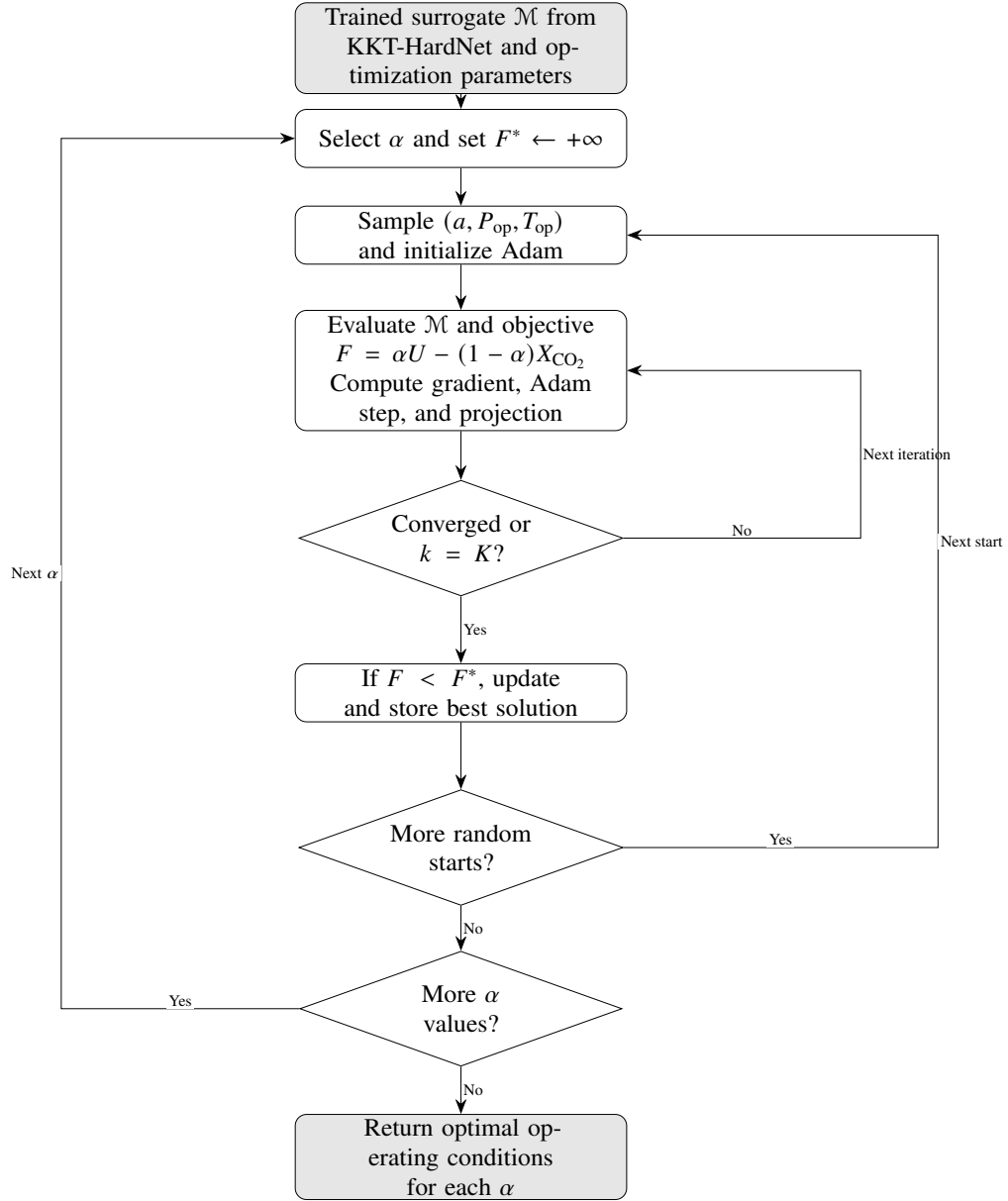

\begin{figure}[!htbp]
    \centering
    \includegraphics[width=0.90\textwidth]{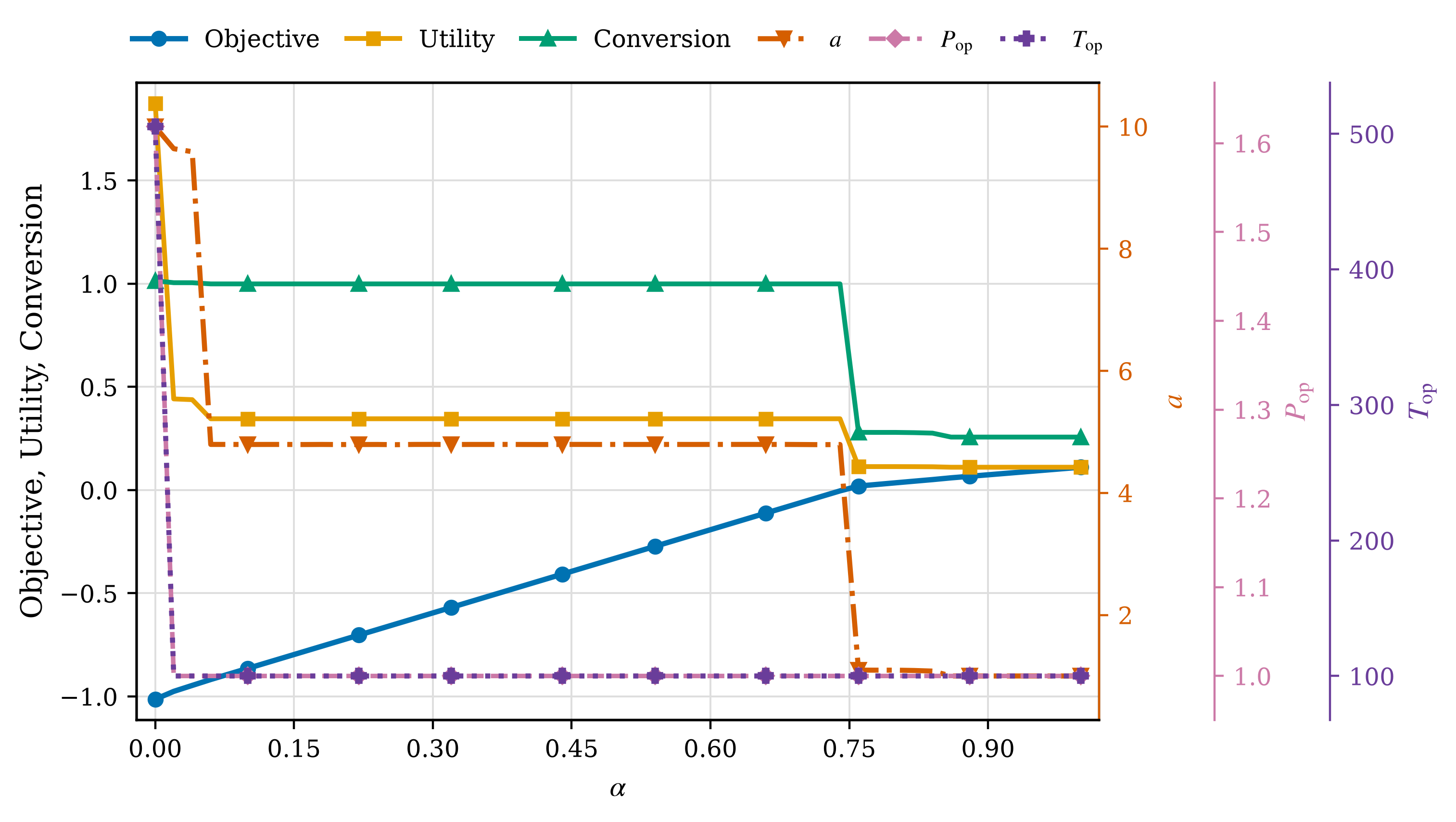}
    \caption{Optimal syngas operating conditions and corresponding performance metrics as a function of the scalarization parameter $\alpha$. The left axis shows the scalarized objective, total utility, and $\mathrm{CO_2}$ conversion, while the three right-side axes show the optimal feed ratio $a$, operating pressure $P_{\mathrm{op}}$, and operating temperature $T_{\mathrm{op}}$.}
    \label{fig:syngas-alpha-operating-conditions}
\end{figure}
\textbf{Optimal operating conditions: }
After training, the KKT-HardNet model is used as a differentiable surrogate to optimize the syngas operating conditions with respect to the feed ratio $a$, operating pressure $P_{\mathrm{op}}$, and operating temperature $T_{\mathrm{op}}$. The scalarized objective is defined as
$F=\alpha U-(1-\alpha)X_{\mathrm{CO_2}}$, where
$X_{\mathrm{CO_2}}=1-F_{\mathrm{CO_2,out}}$ and $\alpha\in[0,1]$
controls the trade-off between utility consumption and $\mathrm{CO_2}$ conversion. The optimization is performed using a multi-start projected Adam scheme in the normalized input space, as shown in Figure~\ref{fig:syngas_projected_adam_flowchart}, with the operating variables constrained to their prescribed bounds. To characterize the trade-off in greater detail, the problem is solved for $\alpha=0,0.02,\ldots,1.0$ using $100$ random initializations for each value of $\alpha$.

The resulting operating conditions and performance metrics are shown in Fig.~\ref{fig:syngas-alpha-operating-conditions}. For small $\alpha$, the optimizer emphasizes conversion, resulting in operating points with nearly complete $\mathrm{CO_2}$ conversion and relatively high utility consumption. Over the broad interval $\alpha\approx0.06$--$0.74$, the optimizer repeatedly identifies an operating condition near $a\approx4.79$, $P_{\mathrm{op}}=1$ atm, and $T_{\mathrm{op}}=100\,^{\circ}\mathrm{C}$, for which the predicted conversion remains close to $0.999$ and the utility is approximately $0.345$. A pronounced transition occurs near $\alpha=0.76$, after which the solution shifts toward lower feed ratio and lower utility consumption, accompanied by a substantial reduction in $\mathrm{CO_2}$ conversion. For $\alpha\gtrsim0.86$, the optimum reaches the lower operating bounds $a=1$, $P_{\mathrm{op}}=1$ atm, and $T_{\mathrm{op}}=100\,^{\circ}\mathrm{C}$, with a predicted utility of approximately $0.110$ and conversion of approximately $0.257$. Since the KKT-HardNet surrogate enforces the elemental-balance constraints by construction, the outer optimization is carried out directly over the bounded operating variables without explicitly reintroducing these balance equations.

%% file: sections/conclusions.tex
\section{Conclusions}\label{conclusions}
Integrating optimization algorithms with machine learning provides a promising route for enforcing physical and operational constraints in data-driven models. In this work, we developed an efficient, scalable, and open-source implementation of KKT-HardNet with modular support for custom neural-network backbones and projection algorithms. Implicit differentiation enables efficient training through iterative projection layers without unrolling the solver iterations.
Across applications in process simulation, real-time optimization and control, and data-driven optimization, KKT-HardNet consistently achieved strong predictive or optimization performance while maintaining tight constraint satisfaction. The results also demonstrate the flexibility of the framework across different problem structures and projection strategies.
Future work will focus on larger-scale applications and problem-specific projection methods that further exploit sparsity and structure. More broadly, we aim to establish KKT-HardNet as a practical development and benchmarking platform for physics-constrained machine learning.


%% file: appendix/appendix_metrics_benchmarks.tex
\section{Benchmarking: metrics and other methods} \label{app:benchmark}
We compare the performance of KKT-HardNet with other existing methods  on the following metrics.
\textbf{Data Loss:} Data loss refers to how close the final prediction is to the available data. We measure the data loss using the standard Mean Squared Error (MSE) metric.

\textbf{Optimality:} Optimality refers to how close the final prediction is to the true optimal solution. This is applicable to all the case studies on optimization.

\textbf{Feasibility:} Feasibility is measured by the constraint violation. The constraint violation is computed as the absolute residual i.e, $\lvert h(x,y)\rvert$ for equality constraint, and $g_+(x,y)$ for inequality where $g_+ = \max(0,g)$. For a feasible point, both the equality and inequality violations should be zero. We evaluate both the maximum violation and the mean violation.

\textbf{Speed:} Speed refers to the time required for inference after the training is finished. The inference time is given for each batch. For optimizer we consider full parallelization and for a fare comparison we divide the total solve time by total number of instances in a batch. For inference, batch size across the case studies remains the same which is $100$ instances.

We compare KKT-HardNet against the following methods:

\begin{itemize}
    \item \textbf{Optimizer:} A traditional optimization solver. We consider IPOPT \cite{wachter2006implementation} as the state-of-the-art optimizer.
    \item \textbf{NN:} A regular artificial neural network model with no soft loss considered. We compare with MLP for the surrogate modeling case studies. 
    \item \textbf{Soft Constrained NN:} We use Physics Informed Neural Networks (PINNs) \cite{raissi2019physics} as the soft constrained methods to compare our model with.
    \item \textbf{ENFORCE:} ENFORCE \cite{lastrucci2025enforce} employs an adaptive depth iterative projection method on the approximated (linearized) constraints. We use the github codespace for ENFORCE including default tunable parameters for comparison.
    \item \textbf{DC3:} The deep constraint completion and correction (DC3) \cite{donti2021dc3} method is state-of-the-art that uses predict and completion method to satisfy the equality constraints. The inequality correction is then done by mapping infeasible points to feasible points using an internal gradient descent step during training. We use the github codespace for DC3 including default tunable parameters for comparison.
\end{itemize}

Across all the case studies we present, a significant portion of the code remains same. These include library imports, backbone setup, training configurations, and solver configurations.

%% file: appendix/appendix_projection_solve.tex
\section{Methods Included to Solve the Projection Problem}\label{app:projection_solution}
We briefly describe reformulation for the KKT conditions to find a feasible solution, a general algorithm to solve the projection optimization problem and the explicit projection for linear equality that we include in our implementation.

\subsection{Explicit Projection: Linear Equality Constraints}
When the constraint space is defined with linear equalities of the form $A\boldsymbol{x}+B\boldsymbol{y}=b$ the KKT condition of the projection in Eq. \ref{eq:projection} yields the analytical solution:

\begin{ceqn}
    \begin{equation}\label{eq:lin-eq}
        \tilde{\boldsymbol{y}} = \hat{\boldsymbol{y}}-B^\top(BB^\top)^{-1}(B\hat{\boldsymbol{y}}+A\boldsymbol{x}-b),
    \end{equation}
\end{ceqn}
which can be written in affine form $\tilde{\boldsymbol{y}} = A^*\boldsymbol{x}+B^*\hat{\boldsymbol{y}}+b^*$ where $A^*=B^\top(BB^\top)^{-1}A$, $B^*=I-B^\top(BB^\top)^{-1}B$, and $b^*=B^\top(BB^\top)^{-1}b$ (see \cite{chen2024physics} for details). This explicit affine form of the solution allows to have a projection with single iteration to ensure feasibility. $A^*$, $B^*$ and $b^*$ can be computed once and used during training and inference making the projection computationally inexpensive.

\subsection{Implicit Projection I: KKT conditions via Fischer-Burmeister Reformulation}
With $\boldsymbol{\lambda}\in\mathbb{R}^p$ and $\boldsymbol{\mu}\in\mathbb{R}^q$ being the Lagrangian multipliers for the equality and inequality constraints, the KKT conditions of the projection in Eq. \ref{eq:projection} for general nonlinear constraints can be written as the following square system of nonlinear equation with slack variables $\boldsymbol{s}\in\mathbb{R}^q$ corresponding to the inequality constraints (see \cite{iftakher2025physics} for details).

\begin{ceqn}
    \begin{subequations}\label{eq:KKT-FB}
        \begin{align}
            \boldsymbol{y} - \hat{\boldsymbol{y}} + \sum_{i=1}^p \lambda_i\nabla_{\boldsymbol{y}}h_i(\boldsymbol{x},\boldsymbol{y}) + \sum_{j=1}^q \mu_j\nabla_{\boldsymbol{y}}g_j(\boldsymbol{x},\boldsymbol{y}) &= \boldsymbol{0},\label{eq:stationary}\\
            h_i(\boldsymbol{x},\boldsymbol{y}) &= 0, \quad && i = 1,2,\dots,p\label{eq:primal-feasibility1}\\ 
            g_j(\boldsymbol{x},\boldsymbol{y}) + s_j & = 0, \quad && j = 1,2,\dots,q\label{eq:primal-feasibility2}\\ 
            \mu_j + s_j - \sqrt{\mu_j^2+s_j^2} & = 0, \quad && j = 1,2, \dots, q. \label{eq:fischer-burmeister},
        \end{align}
    \end{subequations}
\end{ceqn}

where, Eq. \ref{eq:stationary} is the stationarity, \ref{eq:primal-feasibility1} and \ref{eq:primal-feasibility2} are primal feasibility conditions, and Eq. \ref{eq:fischer-burmeister} is the Fischer-Burmeister reformulation of the dual and complementarity conditions without losing any fidelity. Eq. \ref{eq:KKT-FB} can be solved with an iterative Gauss-Newton procedure with Armijo backtracking to find a stationary solution i.e. feasible by construction. It is trivial to show that, if the constraints are only linear equality then Eq. \ref{eq:KKT-FB} yields the same expression as Eq. \ref{eq:lin-eq}. 

\subsection{Implicit Projection II: Primal-Dual Interior Point Method}
Check \cite{wachter2006implementation} for details on the theory and algorithms. We adopt a primitive implementation of the algorithm to fit for batch inference and suitable for GPU runs.

%% file: appendix/appendix_fb_reduced.tex
\section{Reduction and Heuristics on KKT Conditions via Fishcer-Burmeister Reformulation}\label{appendix:fishcer-burmeister-reduced}

In order to solve the KKT system given in Equation \ref{eq:KKT-FB}, a damped Newton method can be applied. We consider $(k)$ as the iteration counter given an iterate ($y_{(k)},\lambda_{(k)},\mu_{(k)},s_{(k)}$). The search direction ($d_{(k)}^y,d_{(k)}^\lambda,d_{(k)}^\mu,d_{(k)}^s$) are obtained from the following linearization:

\begin{ceqn} 
    \begin{equation} \label{eq:lin-kkt-fb}
        \begin{bmatrix}
            H_{(k)}         &   A_{(k)}     &   C_{(k)}     &   0\\
            A_{(k)}^\top    &   0           &   0           &   0\\
            C_{(k)}^\top    &   0           &   0           &   I\\
            0               &   0           &   G_{(k)}^\mu &   G_{(k)}^s
        \end{bmatrix}
        \begin{bmatrix}
            d_{(k)}^y\\
            d_{(k)}^\lambda\\
            d_{(k)}^\mu\\
            d_{(k)}^s
        \end{bmatrix}
        = - \begin{bmatrix}
            y_{(k)}-\hat{y}+A_{(k)}\lambda_{(k)}+C_{(k)}\mu_{(k)}\\
            h(x,y_{(k)})\\
            g(x,y_{(k)})+s_{(k)}\\
            \Phi(\mu_{(k)},s_{(k)})
        \end{bmatrix}.
    \end{equation}
\end{ceqn}

Here, $\Phi(\mu,s)$ is the Fischer-Burmeister equation block, $H_{(k)}:= I + \nabla^2_{yy}[\lambda^\top h(x,y)+\mu^\top g(x,y)]$ is the Hessian of the Lagrangian, $A_{(k)}:=\nabla_y h(x,y)$, $C_{(k)}:=\nabla_y g(x,y)$, $G_{(k)}^\mu=\nabla_\mu \Phi$ and $G_{(k)}^s=\nabla_s\Phi$. Note that, both $G_{(k)}^\mu$ and $G_{(k)}^s$ are diagonal matrix defined as follows:

\begin{ceqn}
    \begin{equation} \label{eq:fb_grad}
        G_{(k)}^\mu = \textrm{diag}(\phi_{\mu_j}); G_{(k)}^s = \textrm{diag}(\phi_{s_j}); \phi_{\mu_j} = 1 - \frac{\mu_j}{\sqrt{\mu_j^2+ s_j^2}}; \phi_{s_j} = 1 - \frac{s_j}{\sqrt{\mu_j^2+ s_j^2}}.
    \end{equation}
\end{ceqn}

While the gradients in Equation \ref{eq:fb_grad} are not defined for $(0,0)$, a smooth Fischer-Burmeister formulation can be used to get the gradient in such cases. Instead of solving the nonsymmetric linear system shown in Equation \ref{eq:lin-kkt-fb} directly, following smaller symmetric system can be solved first,

\begin{ceqn}
    \begin{equation} \label{eq:lin-kkt-fb-reduced}
        \begin{bmatrix}
            H_{(k)}         &   A_{(k)}     &   C_{(k)}\\
            A_{(k)}^\top    &   0           &   0      \\
            C_{(k)}^\top    &   0           &   -{G_{(k)}^s}^{-1}G_{(k)}^\mu
        \end{bmatrix}
        \begin{bmatrix}
            d_{(k)}^y\\
            d_{(k)}^\lambda\\
            d_{(k)}^\mu
        \end{bmatrix}
        = - \begin{bmatrix}
            y_{(k)}-\hat{y}+A_{(k)}\lambda_{(k)}+C_{(k)}\mu_{(k)}\\
            h(x,y_{(k)})\\
            g(x,y_{(k)})+s_{(k)} - {G_{(k)}^s}^{-1}\Phi(\mu_{(k)},s_{(k)})
        \end{bmatrix},
    \end{equation}
\end{ceqn}

with eliminating the last block. The vector $d_{(k)}^s$ can be then obtained from

\begin{ceqn}
    \begin{equation}\label{eq:slack-direction}
        d_{(k)}^s = {G_{(k)}^s}^{-1}\Phi(\mu_{(k)},s_{(k)}) -{G_{(k)}^s}^{-1}G_{(k)}^\mu d_{(k)}^\mu.
    \end{equation}
\end{ceqn}

While, to hold this true ${G_{(k)}^s}$ has to be invertible, which is often time not at the solution, we bypass this with some elementwise diagonal pivoting into the $G_{(k)}^s$ matrix retaining the reduced linear system symmetric. Furthermore, if $A_{(k)}$ is not full rank the system will be singular and a solution might not exist, therefore regularization is used while solving this matrix. After the search direction is obtained the next iterate is obtained with
\begin{ceqn}
    \begin{equation}
        \label{eq:next-step}
        \begin{aligned}
            y_{(k+1)} &= y_{(k)}+\alpha^y_k d_{(k)}^y\\
            \lambda_{(k+1)} &= \lambda_{(k)}+\alpha^\lambda_k d_{(k)}^\lambda\\
            \mu_{(k+1)} &= \mu_{(k)}+\alpha^\mu_k d_{(k)}^\mu\\
            s_{(k+1)} &= s_{(k)}+\alpha^s_k d_{(k)}^s
        \end{aligned}
    \end{equation}
\end{ceqn}

for some $\alpha_{(k)}^y,\alpha_{(k)}^\lambda,\alpha_{(k)}^\mu,\alpha_{(k)}^s\in(0,1]$. In our implementation we kept both methods `\texttt{fischer-burmeister-newton}' and `\texttt{fischer-burmeister-reduced}' with the full linear system and reduced linear system respectively. Moreover, the reduction is only on the slack variables. Hence the performance change is only visible when the constraint set includes inequalities.

%% file: appendix/appendix_control_analysis.tex
\section{Additional Closed-Loop Analysis of the RTO--LMPC Framework}
\label{app:additional-control-analysis}

\subsection{LMPC Control-Effort Analysis}
\label{app:lmpc-control-effort}

\begin{figure}[!htbp]
    \centering
    \includegraphics[width=0.8\textwidth]{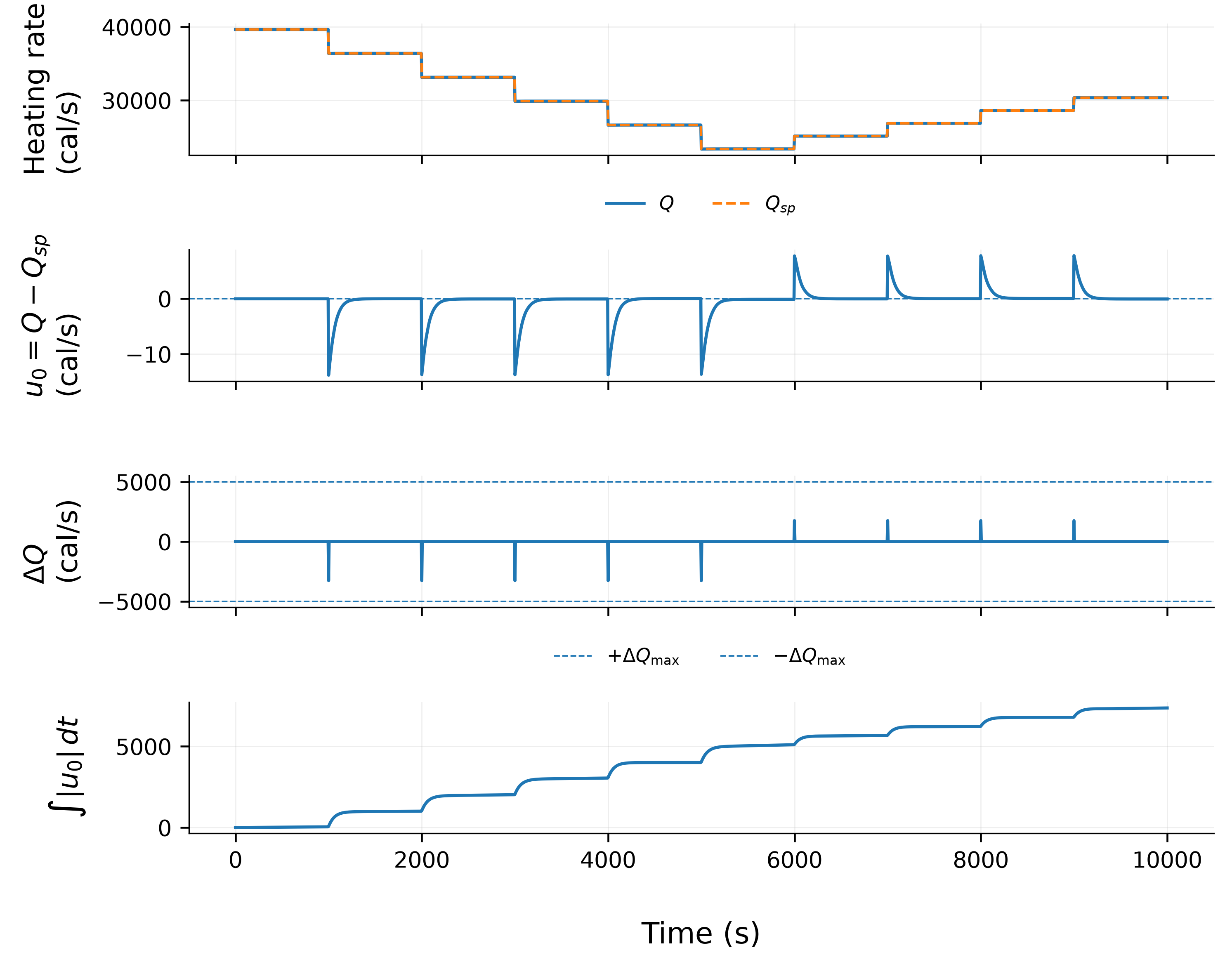}
    \caption{LMPC control-effort analysis during closed-loop RTO--LMPC operation. The first panel compares the applied heat input $Q$ with the RTO steady-state target $Q^{sp}$. The second panel shows the LMPC correction $u_0=Q-Q^{sp}$, which is largest immediately following an RTO set-point update and subsequently decays toward zero. The third panel shows the change between successive implemented heat inputs, with $\Delta Q_{\max}=5\times10^{3}$ cal/s shown as a reference. The final panel shows the cumulative absolute LMPC control effort, $\int |u_0|\,dt$.}
    \label{fig:lmpc-control-effort}
\end{figure}

The LMPC control action is expressed in deviation form as $u_0=Q-Q^{sp}$, allowing the transient corrective action of the controller to be separated from the economically optimal steady-state heat input supplied by the RTO layer. Figure~\ref{fig:lmpc-control-effort} shows that the applied heat input remains very close to $Q^{sp}$ throughout the closed-loop simulation, with appreciable corrective action occurring primarily immediately after an RTO set-point update. The mean absolute control correction was only $0.736$ cal/s, with an RMS value of $2.14$ cal/s and a maximum magnitude of $13.81$ cal/s. This is small relative to the nominal heating rate, which varied between approximately $2.34\times10^{4}$ and $3.96\times10^{4}$ cal/s.

The total variation of the applied heat input was $2.34\times10^{4}$ cal/s, compared with $2.32\times10^{4}$ cal/s for the RTO steady-state heat-input trajectory. Thus, the LMPC correction increased the total control variation by only $198.8$ cal/s, or approximately $0.86\%$, indicating that most of the overall input variation originates from changes in the economic operating target rather than aggressive feedback action. The largest observed change between successive implemented heat inputs was $3.26\times10^{3}$ cal/s, and the applied heat input remained well within its physical bounds throughout the simulation.

\subsection{Closed-Loop Performance under Challenging Heat-Price Profiles}
\label{app:lmpc-price-profiles}

\begin{figure}[!htbp]
    \centering
    \includegraphics[width=0.8\textwidth]{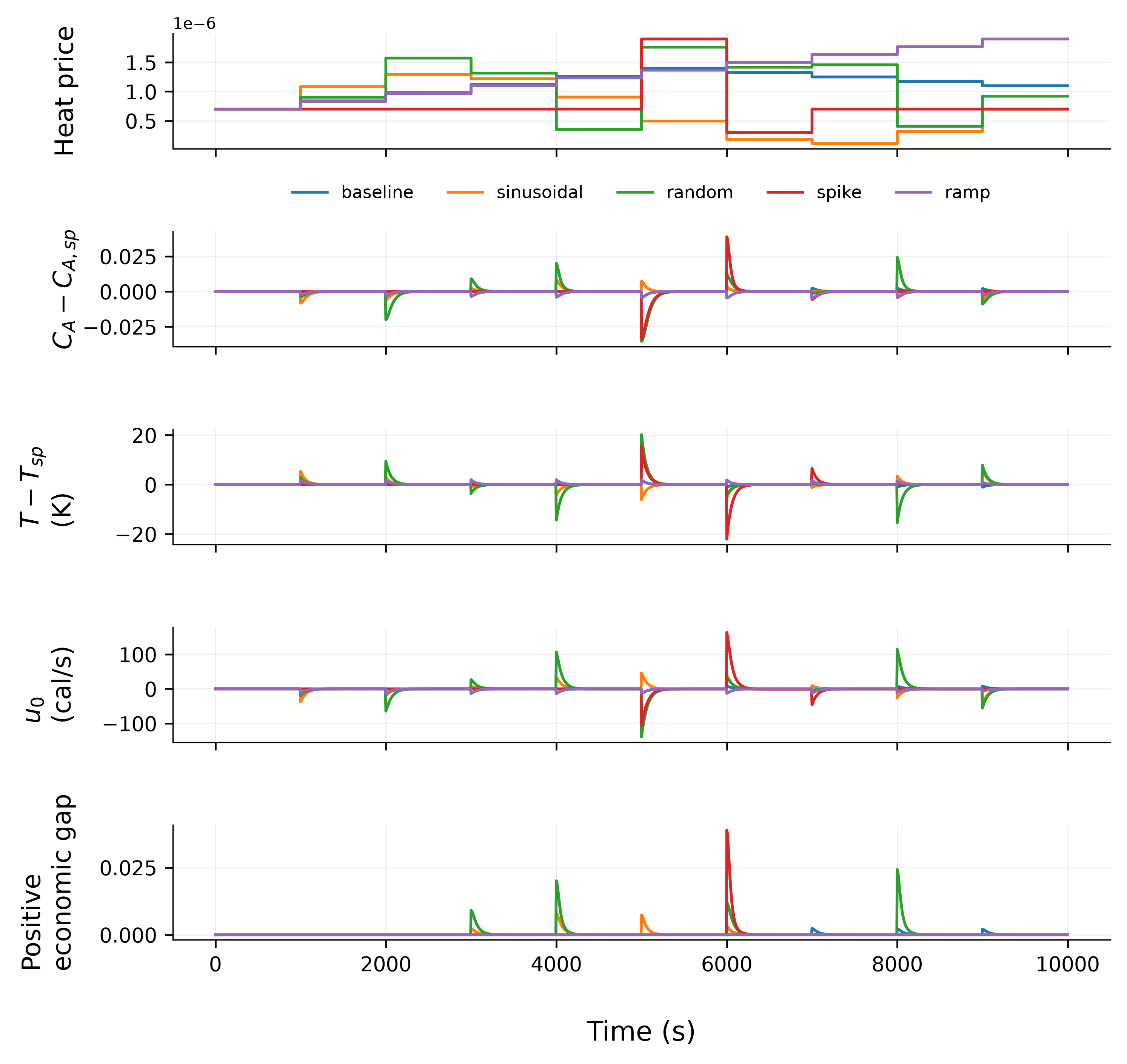}
    \caption{Closed-loop RTO--LMPC performance under challenging heat-price profiles. The first panel shows the imposed baseline, sinusoidal, random, spike, and ramp heat-price trajectories. The subsequent panels show the corresponding $C_A$ tracking error, temperature tracking error, LMPC correction $u_0$, and positive economic gap. More abrupt price changes generate larger transient tracking errors and corrective actions, while all cases remain bounded and satisfy the physical plant limits.}
    \label{fig:lmpc-price-profiles}
\end{figure}

To evaluate the closed-loop response beyond the nominal staircase heat-price trajectory, the coupled RTO--LMPC framework was tested using sinusoidal, random piecewise-constant, sharp-spike, and monotonic-ramp price profiles. Figure~\ref{fig:lmpc-price-profiles} shows that the controller remains stable and bounded for all investigated cases, with zero physical plant-bound violation in every simulation. The baseline and ramp profiles resulted in the smallest tracking errors, with temperature RMSE values of $0.278$ K and $0.299$ K, respectively. More aggressive price variations produced larger transient deviations: the sinusoidal case resulted in a temperature RMSE of $0.761$ K, whereas the random and spike profiles increased the RMSE to $1.85$ K and $1.58$ K, respectively.

\begin{table}[!htbp]
\caption{Closed-loop performance under alternative heat-price profiles.}
\label{tab:lmpc-price-profiles}
\centering
\resizebox{\textwidth}{!}{%
\begin{tabular}{lrrrrrr}
\toprule
Profile & $C_A$ RMSE & $T$ RMSE (K) & Max. $|T-T^{sp}|$ (K) & Max. $|u_0|$ (cal/s) & Integrated positive economic gap & Max. plant violation \\
\midrule
Baseline   & $6.35\times10^{-4}$ & $0.278$ & $1.972$ & $13.81$  & $0.702$ & $0.00$ \\
Sinusoidal & $1.10\times10^{-3}$ & $0.761$ & $6.345$ & $45.66$  & $1.451$ & $0.00$ \\
Random     & $3.87\times10^{-3}$ & $1.845$ & $20.216$ & $139.31$ & $4.238$ & $0.00$ \\
Spike      & $3.41\times10^{-3}$ & $1.580$ & $22.053$ & $163.46$ & $2.149$ & $0.00$ \\
Ramp       & $7.90\times10^{-4}$ & $0.299$ & $1.880$ & $13.16$  & $0.000$ & $0.00$ \\
\bottomrule
\end{tabular}
}
\end{table}

The largest transient errors occurred for the spike and random profiles, for which the maximum temperature tracking errors reached approximately $22.05$ K and $20.22$ K, respectively. These cases also required larger LMPC corrections, with maximum $|u_0|$ values of $163.46$ and $139.31$ cal/s. Despite these more demanding operating scenarios, the process remained within all physical bounds. The integrated positive economic gap increased from $0.702$ for the baseline case to $4.24$ for the random profile, indicating the expected loss in transient economic performance when the RTO target changes more abruptly. In contrast, the smooth ramp profile exhibited essentially zero positive economic gap while maintaining low tracking error.

%% file: appendix/appendix_adam_projection.tex
\section{Multi-Start Projected Adam Optimization}
\label{app:projected-adam}

After training, the KKT-HardNet surrogate is used as a differentiable model
for optimizing the syngas operating conditions. The optimization variables are
the feed-flow ratio, reactor operating pressure, and reactor operating
temperature,
\[
z=(a,P_{\mathrm{op}},T_{\mathrm{op}}).
\]
These variables are restricted to the same operating ranges used to generate
the training data and are normalized to $\mathbf{z}\in[0,1]^3$ during the
optimization.

For a prescribed scalarization parameter $\alpha\in[0,1]$, the objective is
\[
F=\alpha U-(1-\alpha)X_{\mathrm{CO_2}},
\]
where $U$ denotes the total utility requirement and
\[
X_{\mathrm{CO_2}}=1-F_{\mathrm{CO_2,out}}
\]
denotes the $\mathrm{CO_2}$ conversion. The parameter $\alpha$ controls the
trade-off between utility consumption and conversion. Small values of
$\alpha$ place greater emphasis on maximizing conversion, whereas larger
values place greater emphasis on minimizing utility usage.

Because the surrogate model and the resulting scalarized objective are
nonlinear, the optimization problem may contain multiple stationary points.
A single gradient-based run may therefore converge to different local
solutions depending on the initial operating condition. To reduce this
sensitivity to initialization, a multi-start strategy is used. For each value
of $\alpha$, the optimization is repeated from $N_s$ randomly sampled initial
operating points, and the solution with the smallest objective value is
retained. This approach does not guarantee a global optimum, but it provides
a more robust search of the bounded operating domain than a single
initialization.

\begin{algorithm}[!htbp]
\caption{Multi-start projected Adam for syngas operating-condition optimization}
\label{alg:syngas_opt}
\DontPrintSemicolon

\Problem{Trained KKT-HardNet surrogate $\mathcal{M}$}
\Parameter{$\alpha\in[0,1]$, $\eta=0.01$, $\beta_1=0.9$, $\beta_2=0.999$, $\epsilon=10^{-8}$, $K=1000$, $\tau=10^{-8}$, $N_s=100$}

\ForEach{$\alpha\in\{0,0.1,\ldots,1.0\}$}{
    $F^*\leftarrow+\infty$\;

    \For{$s=1,\ldots,N_s$}{
        Sample $z_{k}=(a,P_{\mathrm{op}},T_{\mathrm{op}})$ uniformly within bounds and normalize to $\mathbf{z}\in[0,1]^3$\;

        Initialize $\mathbf{m}_0=\mathbf{0}$ and $\mathbf{v}_0=\mathbf{0}$\;

        \For{$k=1,\ldots,K$}{
            Evaluate $(F_{\mathrm{CO_2,out}},U,\ldots)=\mathcal{M}(z_{k})$\;

            Compute $X_{\mathrm{CO_2}}=1-F_{\mathrm{CO_2,out}}$ and
            $F=\alpha U-(1-\alpha)X_{\mathrm{CO_2}}$\;

            Compute $\mathbf{g}_k=\nabla_{\mathbf{z}}F$\;

            Update
            $\mathbf{m}_k=\beta_1\mathbf{m}_{k-1}+(1-\beta_1)\mathbf{g}_k$
            and
            $\mathbf{v}_k=\beta_2\mathbf{v}_{k-1}+(1-\beta_2)\mathbf{g}_k^2$\;

            Compute
            $\widehat{\mathbf{m}}_k=\mathbf{m}_k/(1-\beta_1^k)$
            and
            $\widehat{\mathbf{v}}_k=\mathbf{v}_k/(1-\beta_2^k)$\;

            Update
            $\widetilde{\mathbf{z}}_{k+1}
            =\mathbf{z}_k-\eta\,
            \widehat{\mathbf{m}}_k/
            (\sqrt{\widehat{\mathbf{v}}_k}+\epsilon)$\;

            Project
            $\mathbf{z}_{k+1}
            =\operatorname{clip}
            (\widetilde{\mathbf{z}}_{k+1},0,1)$\;

            \If{$\|\mathbf{z}_{k+1}-\mathbf{z}_k\|_\infty<\tau$}{
                \textbf{break}\;
            }
        }

        \If{$F<F^*$}{
            $F^*\leftarrow F$\;

            Store
            $(a^*,P_{\mathrm{op}}^*,T_{\mathrm{op}}^*,
            U^*,X_{\mathrm{CO_2}}^*)$\;
        }
    }
}

\Return{Best operating conditions and utility--conversion values for each $\alpha$}
\end{algorithm}

Each local optimization is performed using Adam. At every iteration, the
KKT-HardNet surrogate is evaluated to obtain the process outputs, the
scalarized objective is computed, and its gradient with respect to the
operating variables is obtained. An Adam update is then performed, followed
by projection onto the normalized box $[0,1]^3$. This projection guarantees
that every iterate remains within the prescribed operating bounds and,
therefore, within the domain represented by the surrogate training data.
The optimization terminates when the change in the normalized operating
variables is smaller than the prescribed tolerance or when the maximum number
of iterations is reached.

The complete procedure is summarized in
Algorithm~\ref{alg:syngas_opt}. The multi-start procedure is repeated independently for each value of
$\alpha$, producing a family of operating points that characterizes the
trade-off between $\mathrm{CO_2}$ conversion and utility consumption. The
projection step enforces the operating-variable bounds throughout the
optimization, while the multiple random initializations reduce the
dependence of the final solution on a particular starting point.